\documentclass[10pt,twocolumn,letterpaper]{article}

\usepackage{cvpr}              

\usepackage{float}
\usepackage{pifont}
\usepackage{amssymb}
\usepackage{booktabs} 
\usepackage[linesnumbered,ruled,lined]{algorithm2e}
\usepackage{microtype}
\usepackage{graphicx}
\usepackage{subcaption}
\usepackage{booktabs} 
\usepackage[linesnumbered,ruled,lined]{algorithm2e}
\usepackage{multirow} 
\usepackage{makecell}
\usepackage{colortbl}
\usepackage{color}
\usepackage{pifont}
\usepackage{amssymb}
\newtheorem{lemma}{Lemma}
\newtheorem{theorem}{Theorem}

\newcommand{\cmark}{\color{green}\ding{51}}%
\newcommand{\xmark}{\color{red}\ding{55}}

\definecolor{cvprblue}{rgb}{0.21,0.49,0.74}
\usepackage[pagebackref,breaklinks,colorlinks,allcolors=cvprblue]{hyperref}

\def\paperID{6340} 
\def\confName{CVPR}
\def\confYear{2025}

\title{OODD: Test-time Out-of-Distribution Detection with Dynamic Dictionary}

\author{
  Yifeng Yang$^{1}$, Lin Zhu$^{1}$, Zewen Sun$^{2}$, Hengyu Liu$^{3}$, Qinying Gu$^{4}$, Nanyang Ye$^{1}$\thanks{Nanyang Ye is the corresponding author.} \\
  $^{1}$ Shanghai Jiao Tong University \quad $^{2}$ Tianjin University\\
  $^{3}$ The Chinese University of Hong Kong \quad $^{4}$ Shanghai Artificial Intelligence Laboratory \\
  {\tt\small \{zxk1212, zhulin\_sjtu, ynylincoln\}@sjtu.edu.cn}, 
  {\tt\small 3022244294@tju.edu.cn}, \\
  {\tt\small piang.lhy@link.cuhk.edu.hk}, 
  {\tt\small guqinying@pjlab.org.cn}
}

\newcommand{\calX}{\mathcal{X}}
\newcommand{\calY}{\mathcal{Y}}

\newcommand{\bx}{\mathbf{x}}
\newcommand{\bz}{\mathbf{z}}

\newcommand{\triangleitem}{\item[\(\triangleright\)]}

\newcolumntype{L}[1]{>{\raggedright\arraybackslash}p{#1}}
\newcolumntype{C}[1]{>{\centering\arraybackslash}p{#1}}
\newcolumntype{R}[1]{>{\raggedleft\arraybackslash}p{#1}}
\definecolor{lightgreen}{RGB}{76,175,80}

\begin{document}

\maketitle

\begin{abstract}
Out-of-distribution (OOD) detection remains challenging for deep learning models, particularly when test-time OOD samples differ significantly from training outliers. We propose \textbf{OODD}, a novel test-time OOD detection method that dynamically maintains and updates an OOD dictionary without fine-tuning. Our approach leverages a priority queue-based dictionary that accumulates representative OOD features during testing, combined with an informative inlier sampling strategy for in-distribution (ID) samples. To ensure stable performance during early testing, we propose a dual OOD stabilization mechanism that leverages strategically generated outliers derived from ID data. To our best knowledge, extensive experiments on the OpenOOD benchmark demonstrate that OODD significantly outperforms existing methods, achieving a 26.0\% improvement in FPR95 on CIFAR-100 Far OOD detection compared to the state-of-the-art approach. Furthermore, we present an optimized variant of the KNN-based OOD detection framework that achieves a 3x speedup while maintaining detection performance. Our code is available at \url{https://github.com/zxk1212/OODD}.
\end{abstract}

\section{Introduction}
Deep learning has demonstrated impressive capabilities but often exhibits unpredictable behavior when faced with unknown situations, such as receiving data unrelated to its training tasks. What's worse, the model tends to incorrectly classify unknown out-of-distribution (OOD) data into one of the in-distribution (ID) classes with high confidence \cite{goodfellow2014explaining, nguyen2015deep}.
Existing solutions in OOD detection areas can be broadly categorized into three main types: (i) post-hoc methods, (ii) fine-tuning methods, and (iii) test-time calibrated methods.

Post-hoc methods typically design an OOD score function to analyze model outputs (e.g., logits), features, or layer-wise statistics \cite{hendrycks2016baseline, liang2017enhancing, lee2018simple, liu2020energy, sun2022out, djurisic2022extremely}. During inference, these methods discriminate between ID and OOD samples based on the OOD scores of test samples.
In contrast, fine-tuning methods
\cite{hendrycks2018deep, chen2021atom, zhang2023mixture, zhu2023diversified} require fine-tuning the model that has been pre-trained on ID data, using a combination of an auxiliary dataset with outliers and ID samples. These methods are often called outlier exposure (OE) based methods, aiming to make the classifier outputs more uniformly distributed when exposed to outliers.
However, when the test-time OOD samples differ significantly from the training outliers, the model may still assign high confidence to test-time OOD samples \cite{wang2023out}. While recent works \cite{yang2023auto,fan2024test,anonymous2024TULIP} attempt to calibrate OOD detection during test time, their batch-wise parameter updates can lead to catastrophic forgetting, unstable detection performance, and even impact ID classification. 
We consider that the key to test-time OOD detection is utilizing the latent OOD features. Particularly in the medical field, when dealing with unknown diseases during the early stages of an epidemic, there is an urgent need to specifically collect the features of these unknown diseases \cite{gonzalez2022distance}.

\begin{figure}[t]
  \centering
  \includegraphics[scale=0.27]{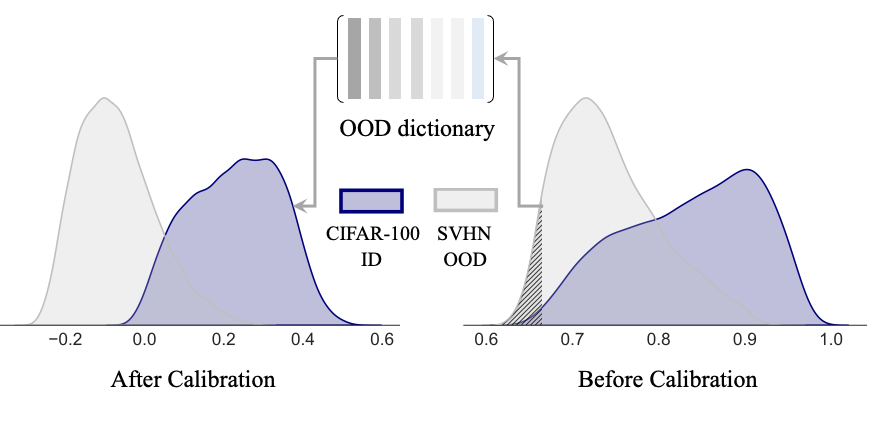}
  \vspace{-6mm}
  \caption{
Before calibration, we dynamically feed the lower tail of the OOD score distribution into an OOD dictionary through a priority queue and then use this dictionary to calibrate the OOD scores.
  }
  \label{fig:intro}
  \vspace{-4mm}
\end{figure}

Given this, we propose a method called \emph{\textbf{OODD} (test-time \textbf{O}ut-\textbf{O}f-\textbf{D}istribution detection with dynamic \textbf{D}ictionary)}, which does not impose any burden during the training phase, nor does it require optimization or fine-tuning at test time. It only requires the addition of a dynamic OOD dictionary managed by a priority queue, enabling calibration at test time with just a single matrix multiplication with the keys in the OOD dictionary. The intuition behind our method is illustrated in Figure \ref{fig:intro}.
Specifically, based on the OOD scores at test time, we collect features of OOD samples in real time, focusing on those with lower scores in the left tail of the OOD score distribution. These features are then stored in the dynamic OOD dictionary by a priority queue. During testing, we calculate the similarity between the features of the test sample and the features stored in the dynamic OOD dictionary. Then, according to the similarity between the sample’s feature and the features stored in the dynamic OOD dictionary, we can adaptively calibrate the OOD score. 
Additionally, by using a priority queue to manage the features in the OOD dictionary, our approach mitigates catastrophic forgetting and detection fluctuation in other methods by updating parameters with minibatches. 
Notably, we designed a strategy to initialize this priority queue with outliers generated from the training ID data, alleviating the detection fluctuations caused by inaccurate estimates during the early testing phase.
To evaluate the effectiveness of our method, we conducted comprehensive experiments on the OpenOOD benchmark \cite{zhang2023openood}, which is one of the authoritative platforms for testing OOD detection methods. 

Extensive experiments confirm that our method can effectively calibrate the OOD score during testing. Compared to the most relevant baseline TULIP \cite{anonymous2024TULIP}, our method significantly improves the FPR95 from 58.17\% to 24.74\% on CIFAR-100 Far OOD. Our key contributions are as follows:
\begin{itemize}
	\item We propose a novel test-time OOD detection method based on a dynamic OOD dictionary. Unlike previous approaches, it doesn’t require prior outlier samples from auxiliary datasets. Additionally, we devise a specific initialization for the OOD dictionary to alleviate the initial test-time fluctuations.
	\item Our approach achieves state-of-the-art performance on the OpenOOD benchmark, surpassing current baselines. Moreover, it demonstrates strong complementarity with existing post-hoc methods, effectively enhancing their performance when used in combination.
 	\item	To optimize computational efficiency, we use cosine similarity to replace the Euclidean distance used in KNN-based OOD detection methods without comprising performance. We provide a rigorous theoretical analysis establishing the relationship between these distance metrics and justify their equivalence in discriminative ability for OOD detection.
\end{itemize}

\begin{figure*}[ht]
\begin{center}
  \includegraphics[scale=0.405]{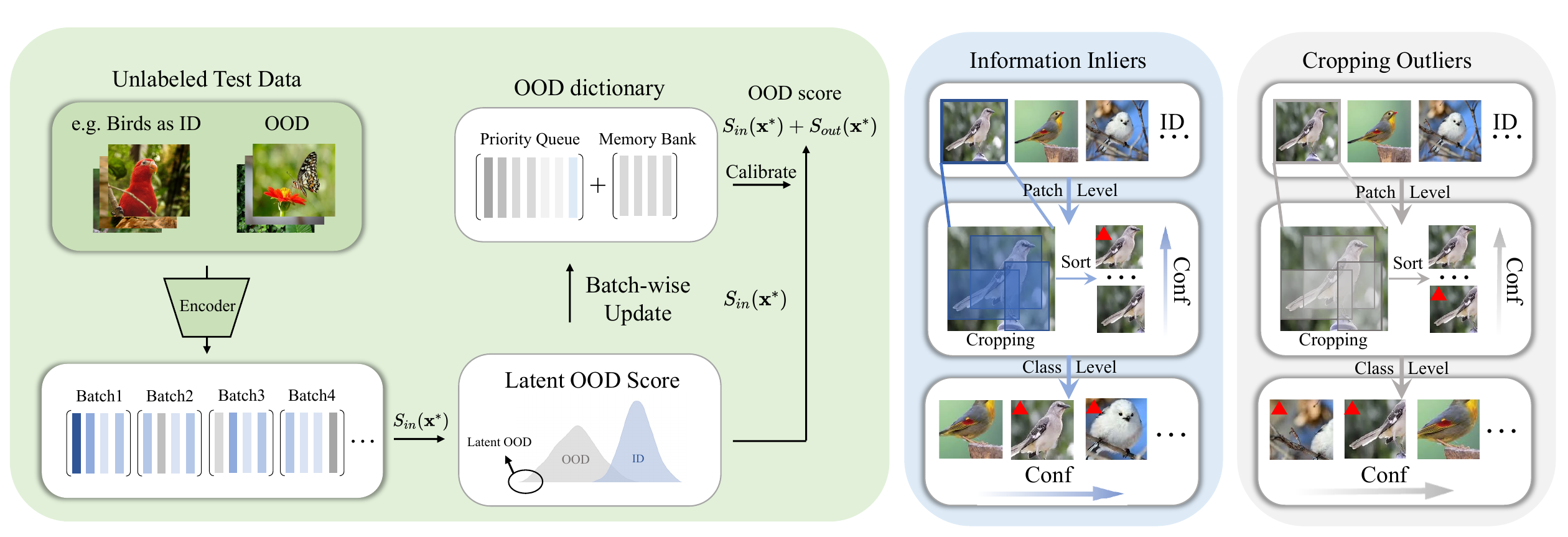}
  
\caption{
Overview of the proposed OODD framework, which performs real-time, batch-wise updates to the OOD dictionary using a priority queue based on the left-tail distribution of OOD score. Using training ID samples, we generate multiple random crops for each sample \cite{sun2024diversity}, which are then filtered at the patch and class levels. High-confidence inliers are selected to compute the latent OOD score, while low-confidence outliers initialize the priority queue, allowing for a more robust and adaptive OOD detection process.
}
\label{fig:ov}
\end{center}
\vspace{-0.3cm}
\end{figure*}

\section{Related Works}
\textbf{Learning from auxiliary data.}
According to auxiliary data type, learning from auxiliary data methods can be classified into two categories: active learning and outlier exposure. In the first category, 
Scone \cite{bai2023feed} addresses OOD detection by utilizing auxiliary data consisting of a mixture of ID and OOD samples. They assign pseudo-labels to this auxiliary data, enabling the model to adapt to these pseudo-labeled samples.  Other notable works in this category include AL-OOD \cite{schmidt2024unified}, DUL \cite{zhang2024best}.
The second category, known as Outlier Exposure (OE) \cite{hendrycks2018deep}, uses auxiliary data where the class space usually does not overlap with ID or test-time OOD samples. By leveraging such data, the aim is to make the model more robust against test-time OOD samples \cite{hendrycks2018deep, chen2021atom, zhang2023mixture, zhu2023diversified}.

\noindent\textbf{Test-Time Adaptation.}
Test-time adaptation methods can generally be categorized as:
(i) Test-Time Domain Adaptation (TTDA):
In this situation, we can obtain samples of unlabeled data in test domain once before deployment \cite{shu2022test}. However, for OOD detection scenarios, it's difficult to gather all potential test domain data beforehand, particularly in the dynamic open world.
(ii) Test-Time Batch Adaptation (TTBA): the model undergoes adaptive updates each time it processes a new batch, helping it handle distributional shifts over time. RTL++ \cite{fan2024test} assumes a linear correlation between OOD scores and features, but the feature distribution overlap between hard OOD and ID samples challenges this simple assumption.  AUTO \cite{yang2023auto} directly modifies the model at test time using stochastic gradient descent to produce low-confidence predictions for potential OOD samples. Our work also belongs to this category, but we don't require strong assumptions or fine-tuning.
(iii) Test-Time Prior Adaptation (TTPA):
This adjusts prior information within the model, such as class or feature distribution at test time. For example, TUIP \cite{anonymous2024TULIP} introduces priors for ID and OOD samples at test time to calibrate model uncertainty.

\noindent\textbf{Deep K-Nearest Neighbor for OOD detection.} \label{sec:OE}
\cite{sun2022out} proposes using K-nearest neighbors (KNN) for OOD detection. The approach first normalizes the penultimate layer features of a neural network and computes the Euclidean distance between a test sample's feature and all training sample features. The opposite of the distance to the K-th nearest neighbor is then used as the OOD score of this test sample. Despite its simplicity, this KNN-based method nearly outperforms other post-hoc methods and even fine-tuning methods in OpenOOD benchmark \cite{zhang2023openood}.

\section{Preliminaries}
\textbf{Setup.} We consider a supervised multi-class classification setting where the input space is denoted by $\mathcal{X}$ and the label space by $\mathcal{Y} = \{1, 2, …, C\}$. The training set $ \{(\bx_i, y_i)\}_{i=1}^n$ is drawn \emph{i.i.d.} from the joint distribution $P_{\mathcal{X}\mathcal{Y}}$. Let $f: \calX \mapsto \mathbb{R}^{|\calY|}$ be a model parameterized by $\theta$, trained on samples drawn from $P_{\mathcal{X}\mathcal{Y}}$, which outputs logits.
Then the logits are transformed by a softmax layer: $\hat{p_k}=e^{f_k} / \sum_{j=1}^C e^{f_j}$ for $k \in \mathcal{Y}$, where the class probability of a sample $\bx$ is defined as $\mathrm{P}\left(y=k \mid \theta, \bx\right)=\hat{p_k}$, and the confidence is defined as $\operatorname{Conf}\left(\bx\right)=\max _y \mathrm{P}\left(y \mid \theta, \bx\right)$.
Importantly, we could use the latent feature $\bz= \phi(\bx) $ for OOD detection, where $\phi: \mathcal{X} \mapsto \mathbb{R}^{d}$ is an image encoder, and $d$ is the dimensionality of the latent space.

\noindent\textbf{Problem Statement.} At test time, our goal is to find a good OOD detector, which can distinguish whether a sample $\bx \in \mathcal{X}$ is from ID or not OOD. The decision function $D_{\tau}$ is usually defined as follows:
\vspace{-0.1cm}
\begin{equation}
  D_{\tau}(\mathbf{x})=\left\{\begin{array}{ll}
  ID  & S\left(\mathbf{x}\right) \geq \tau \\
  OOD & S\left(\mathbf{x}\right)<\tau
  \end{array},\right.
  \end{equation}
where samples with lower OOD scores $S(\bx)$ are classified as OOD and vice versa, and  $\tau$ is a handcraft threshold, which is commonly chosen so that a high fraction (e.g., 95\%) of ID data is correctly distinguished. 

\section{Methodology}
In this section, we present our proposed method for informative inlier and dynamic OOD dictionary construction. We first describe the process of informative inlier sampling (IIS) in Sec \ref{iis}, where key patches with high confidence are selected to create a representative ID dictionary. In Sec \ref{dood}, we extend the KNN-based method to build a dynamic OOD dictionary using a priority queue that is progressively updated with latent OOD samples to calibrate OOD score. To stabilize the OOD detection process, we introduce a Dual OOD Stabilization (DOS) strategy. Finally, we propose three approaches for obtaining outliers, which are crucial for initializing the OOD dictionary and stabilizing the detection process. An overview of the proposed method is shown in Figure \ref{fig:ov}.

\subsection{Informative Inlier Sampling} \label{iis}
Unlike the KNN-based OOD detection \cite{sun2022out}, which uses all or randomly selected training data as an ID dictionary to compute the similarity between the query (latent features) of a test sample, we propose that the ID dictionary should be more informative and representative.
To sample the informative inliers from training ID dataset, we capture the key patch of samples with high OOD scores at the patch level and the class level respectively.
Specifically, to sufficiently explore the vicinal space of training samples, we perform multiple random cropping \cite{sun2024diversity} on each training ID sample $\bx_i$ to obtain the set $X_i^{crop}=\left\{\bx_{i, m}^{crop}\right\}_{m=1}^M$, where $M$ is the number of random cropping. Then, based on the confidence of each patch $\operatorname{Conf}(\mathbf{x}^{crop}_{i, m})$ for each sample, we select the patch with the highest confidence. After selecting the highest-confidence patches for all samples, we further choose the top $\alpha\%$ of patches based on their confidence within each class.
The latent features of the final selected patches are then saved in ID dictionary, denoted as $\mathcal{K}^{id}_{n^{\prime}} = \{k_1^{id}, k_2^{id}, \dots, k^{id}_{n^{\prime}}\}$, where $n^{\prime}=[\alpha\% \cdot n]$, $n$ is the number of training samples and $[\cdot]$ denotes the rounding function.

\subsection{Dynamic OOD Dictionary} \label{dood}
In a similar way to constructing the ID dictionary, we can easily extend this approach to construct the OOD dictionary, but the OOD dictionary should be dynamic in
the sense that OOD samples randomly appear and are sampled as keys by latent OOD scores $S_{in}(\bx)$ during testing. Our hypothesis is that
good OOD latent features can be saved in the OOD dictionary that covers a rich set of OOD samples, while the OOD
dictionary keys are kept as consistent as possible despite its
evolution. 

\noindent\textbf{Latent OOD Score.}
we derive the query $q (\bx^*) = \Phi(\bx^*)$ for a test sample $\bx^*$ and compute the cosine similarity $\cos (k^{id}_i, q)$ with each key $k^{id}_i$ in ID dictionary $\mathcal{K}^{id}_{n^{\prime}}$. Then, we denote the sorted list of these similarities in descending order as \(\cos(k^{id}_{(1)}, q) \geq \cos(k^{id}_{(2)}, q) \geq \cdots \geq \cos(k^{id}_{(n^{\prime})}, q)\).
Finally, the latent OOD score is given by:
\begin{equation}
  S_{in}(\bx^*) =  \cos(k^{id}_{(\mathbb{K})}, q)
  \label{eq:s_in}
  \end{equation}
Where the $\mathbb{K}$-th largest cosine similarity serves as the latent OOD score \( S_{in}(\bx^*) \) for the test-time sample.

\noindent\textbf{OOD dictionary as a priority queue.}
During test time, the core of our motivation is maintaining the OOD dictionary as a priority queue of potential OOD sample keys (or latent features). This configuration enables the reuse of encoded keys from previous mini-batches by leveraging a priority queue structure that separates the dictionary size from the mini-batch size. Consequently, the OOD dictionary can be much larger than the mini-batch, and the OOD dictionary size can be easily adjusted as a hyperparameter.
Dictionary samples are progressively updated, with lower potential OOD scores (i.e., more likely to be OOD) replacing existing samples. The priority queue maintains the sample with the highest OOD score at the front. A new sample from the mini-batch is enqueued only if it has a lower OOD score than the queue front, and the front needs to be dequeued first when the dictionary is full. Further, we denote this OOD dictionary as $\mathcal{K}^{ood}_{l} = \{ k_1^{ood}, k_2^{ooid}, \dots, k^{ood}_{l^{\prime}} \}$ where $l$ represents the size of the proposed priority queue and $l \geq l^{\prime}$.
Thus, the OOD dictionary $\mathcal{K}^{ood}_{l}$ represents a dynamically sampled latent OOD subset of the test-time data, and keeping the OOD dictionary up to date incurs little computational cost. 

\noindent\textbf{Dual OOD Stabilization.}
Although subsequent experiments show that initializing the OOD dictionary as empty has little impact on OOD detection results in most scenarios, the empty OOD dictionary may lead to highly unstable scores during the early stage. To maintain historical information inertia, we propose the Dual OOD Stabilization (DOS) strategy to mitigate substantial fluctuations in calibrated OOD scores caused by dramatic changes in latent OOD data within the priority queue. Specifically, we utilize some outliers (with their obtained approaches further detailed later) as a memory bank and others to initialize the priority queue. Under this setup, the OOD dictionary encompasses not only the priority queue but also the memory bank, which can be represented as:
$\mathcal{K}^{ood}_{total} = \mathcal{K}^{ood}_l \cup \mathcal{K}^{ood}_{mb} $
where $\mathcal{K}^{ood}_{mb}$ represents samples in the memory bank, and the memory bank size is denoted as $mb$. Similar to Equation \ref{eq:s_in}, we can obtain the negative cosine similarity between the query and each of the keys in the OOD dictionary during testing, as follows:

\begin{equation}
  S_{out}(\bx^*) =  -\cos(k^{ood}_{(\mathbb{\hat{K}})}, q)
  \end{equation}
  where $\mathbb{\hat{K}}$-th largest cosine similarity serves as the calibrated OOD score \( S_{out}(\bx^*) \) for the test-time sample. 
Finally,the OOD score in our method is integrated as follows:
\begin{equation}
  S(\bx^*) = S_{in}(\bx^*)+S_{out}(\bx^*)
  \end{equation}

However, a critical yet fully underexplored
question thus arises: 
\begin{center}
  \emph{How do we obtain these outliers?}
\end{center}
We explore three approaches for obtaining outliers in our subsequent experiments:
\begin{itemize}[leftmargin=*]
\triangleitem \textbf{C-Out (Cropping Outliers):} Use multiple random cropping augmentation \cite{sun2024diversity} on the training ID data to generate low-confidence outliers that originate from ID data.
\triangleitem \textbf{T-Out (Target Outliers):} Introduce a small amount of data that shares the same distribution as the test-time OOD data as outliers.
\triangleitem \textbf{D-Out (Different test-time Outliers)}: This type of outlier is derived from a distribution different from both the ID and test-time OOD data.
\end{itemize}
Notably, our proposed C-Out method does not introduce any prior knowledge beyond training ID. The process it uses to generate outliers is the reverse of what is described in Sec \ref{iis}. In contrast, T-Out incorporates priors aligned with the test-time OOD distribution, while D-Out needs to leverage samples from external datasets. Subsequent experiments demonstrate that C-Out alone can achieve strong performance, even without using priors beyond the training ID data.

\section{Experiments}

\begin{table*}[!ht]
      \centering
\caption{Comparison with state-of-the-art methods on the CIFAR benchmarks, where the best results are in bold.}
      \vspace{-0.4cm}
      \vskip0.1in
      \resizebox{0.87\linewidth}{!}{
       \begin{tabular}
 {C{2.75cm}C{2.75cm}|C{2.75cm}|C{2.75cm}C{2.75cm}|C{2.75cm}C{2.75cm}}
       
        \toprule
        \multirow{2}[2]{*}{\makecell{Dataset}} & \multicolumn{1}{c|}{\multirow{2}[2]{*}{Method}} & {\multirow{2}[2]{*}{ID ACC↑}} &  \multicolumn{2}{c|}{Near OOD}  & \multicolumn{2}{c}{Far OOD} \\
        & \multicolumn{1}{c|}{} & \multicolumn{1}{c|}{}   & \multicolumn{1}{c}{AUROC↑} &  FPR95↓  & \multicolumn{1}{c}{AUROC↑} &  FPR95↓ \\
        \midrule
        \multicolumn{1}{c}{\multirow{8}[0]{*}{\makecell{CIFAR-10}}} 
        & MaxLogits \cite{hendrycks2019scaling}& \textbf{95.06 ± 0.30} & 87.52 ± 0.47 & 61.32 ± 4.62 & 91.10 ± 0.89 & 41.68 ± 5.27 \\
        & ODIN \cite{liang2017enhancing} & \textbf{95.06 ± 0.30}  & 82.87 ± 1.85 & 76.19 ± 6.08 & 87.96 ± 0.61 & 57.62 ± 4.24 \\
        & MD \cite{lee2018simple} & \textbf{95.06 ± 0.30} & 84.20 ± 2.40 & 49.90 ± 3.98 & 89.72 ± 1.36 & 32.22 ± 3.40 \\
        & Energy \cite{liu2020energy} & \textbf{95.06 ± 0.30} & 87.58 ± 0.46 & 61.34 ± 4.63 & 91.21 ± 0.92 & 41.69 ± 5.32 \\
        
        & KNN \cite{sun2022out} & \textbf{95.06 ± 0.30} & 90.64 ± 0.20 & 34.01 ± 0.38 & 92.96 ± 0.14 & 24.27 ± 0.40 \\
        & ViM \cite{djurisic2022extremely} & \textbf{95.06 ± 0.30} & 88.68 ± 0.28 & 44.84 ± 2.31 & 93.48 ± 0.24 & 25.05 ± 0.52 \\
        & RTL++ \cite{fan2024test} & \textbf{95.06 ± 0.30} & 88.76 ± 0.01 &54.03 ± 0.21 &87.06 ± 0.07 & 64.30 ± 0.21 \\
        & TULIP \cite{anonymous2024TULIP} & \textbf{95.06 ± 0.30} & 89.67 ± 0.24 &  \textbf{33.80 ± 0.59} & 92.55 ± 0.13 &  24.43 ± 0.17\\
        \rowcolor{gray!25}&  OODD & \textbf{95.06 ± 0.30} & \textbf{90.96 ± 0.21} & 36.01 ± 0.44 & \textbf{95.77 ± 0.12} & \textbf{17.44 ± 0.44} \\

              \midrule
              \midrule
              \multicolumn{1}{c}{\multirow{8}[0]{*}{\makecell{CIFAR-100}}} 
              & MaxLogits \cite{hendrycks2019scaling}& \textbf{77.25 ± 0.10} & 81.05 ± 0.07 & 55.47 ± 0.66 & 79.67 ± 0.57 & 56.73 ± 1.33 \\
              & ODIN \cite{liang2017enhancing} & \textbf{77.25 ± 0.10}  & 79.90 ± 0.11 & 57.91 ± 0.51 & 79.28 ± 0.21 & 58.86 ± 0.79 \\
              & MD \cite{lee2018simple} & \textbf{77.25 ± 0.10} & 58.69 ± 0.09 & 83.53 ± 0.60 & 69.39 ± 1.39 & 72.26 ± 1.56 \\
              & Energy \cite{liu2020energy} & \textbf{77.25 ± 0.10} & 80.91 ± 0.08 & 55.62 ± 0.61 & 79.77 ± 0.61 & 56.59 ± 1.38 \\
              & KNN \cite{sun2022out} & \textbf{77.25 ± 0.10} & 80.18 ± 0.15 & 61.22 ± 0.14 & 82.40 ± 0.17 & 53.65 ± 0.28 \\
              & ViM \cite{djurisic2022extremely} & \textbf{77.25 ± 0.10} & 74.98 ± 0.13 & 62.63 ± 0.27 & 81.70 ± 0.62 & 50.74 ± 1.00 \\
              & RTL++ \cite{fan2024test} & \textbf{77.25 ± 0.10} &79.24 ± 0.03 &63.92 ± 0.27 & 80.76 ± 0.03& 67.14 ± 0.07 \\
              & TULIP \cite{anonymous2024TULIP} & \textbf{77.25 ± 0.10} & 81.29 ± 0.26  & \textbf{55.07 ± 0.73} & 79.63 ± 0.94 & 58.17 ± 1.78  \\
              \rowcolor{gray!25}&  OODD & \textbf{77.25 ± 0.10} & \textbf{82.10 ± 0.20} & 56.98 ± 1.09 & \textbf{93.64 ± 0.31} & \textbf{24.74 ± 1.30} \\

               \bottomrule
        \end{tabular}%
        }
        
      \label{tab:main_result-cifar}%
    \end{table*}%

    \begin{table*}[!ht]
      \centering
\caption{Comparison with state-of-the-art methods on the ImageNet benchmarks, where the best results are in bold.}
      
      \vskip0.1in
      \vspace{-0.4cm}
      \resizebox{0.87\linewidth}{!}{
       \begin{tabular}
 {C{2.75cm}C{2.75cm}|C{2.75cm}|C{2.75cm}C{2.75cm}|C{2.75cm}C{2.75cm}}
       
        \toprule
        \multirow{2}[2]{*}{\makecell{Dataset}} & \multicolumn{1}{c|}{\multirow{2}[2]{*}{Method}} & {\multirow{2}[2]{*}{ID ACC↑}} &  \multicolumn{2}{c|}{Near OOD}  & \multicolumn{2}{c}{Far OOD} \\
        & \multicolumn{1}{c|}{} & \multicolumn{1}{c|}{}   & \multicolumn{1}{c}{AUROC↑} &  FPR95↓  & \multicolumn{1}{c}{AUROC↑} &  FPR95↓ \\
        \hline
       \multicolumn{1}{c}{\multirow{8}[0]{*}{\makecell{ImageNet-200}}} 
              & MaxLogits \cite{hendrycks2019scaling}& \textbf{86.37± 0.08} &82.90 ± 0.04& 59.76 ± 0.59 & 91.11 ± 0.19 & 34.03 ± 1.21 \\
              & ODIN \cite{liang2017enhancing} & \textbf{86.37 ± 0.08}  &  80.27± 0.08 &  66.76 ± 0.26 & 91.71 ± 0.19 & 34.23 ± 1.05\\
              & MD \cite{lee2018simple} & \textbf{86.37± 0.08} & 61.93 ± 0.51 & 79.11 ± 0.31 & 74.72 ± 0.26
              & 61.66 ± 0.27 \\
              & Energy \cite{liu2020energy} & \textbf{86.37± 0.08} & 82.50 ± 0.05 & 60.24 ± 0.57 & 90.86 ± 0.21  &  34.86 ± 1.30 \\
              
              & KNN \cite{sun2022out} & \textbf{86.37± 0.08} & 81.57 ± 0.17 & 60.18 ± 0.52 & 93.16 ± 0.22 & 27.27 ± 0.75 \\
              & ViM \cite{djurisic2022extremely} & \textbf{86.37± 0.08} & 78.68 ± 0.24 & 59.19 ± 0.71 & 91.26 ± 0.19 & 27.20 ± 0.30 \\
              & RTL++ \cite{fan2024test} & \textbf{86.37± 0.08} & 81.02 ± 0.08 & 63.99 ± 0.29&93.11 ± 0.02 & 30.38 ± 0.10 \\
              & TULIP \cite{anonymous2024TULIP} & \textbf{86.37± 0.08} & 83.77 ± 0.06 & 54.51 ± 0.35 & 91.03 ± 0.09 & 33.94 ± 0.51 \\
              \rowcolor{gray!25}&  OODD & \textbf{86.37± 0.08} & \textbf{85.74 ± 0.03} & \textbf{53.70 ± 0.14} & \textbf{95.47 ± 0.03} & \textbf{20.89 ± 0.12} \\
              \midrule
              \midrule
              
              \multicolumn{1}{c}{\multirow{8}[0]{*}{\makecell{ImageNet-1K}}} 
              & MaxLogits \cite{hendrycks2019scaling}& \textbf{76.18} & 76.46 & 67.82 & 89.57 & 38.22 \\
              & ODIN \cite{liang2017enhancing} & \textbf{76.18}  &  74.75 &  72.50 & 89.47 & 43.96\\
              & MD \cite{lee2018simple} & \textbf{76.18} & 55.44 & 85.45 & 74.25
              & 62.92 \\
              & Energy \cite{liu2020energy} & \textbf{76.18} & 75.89 & 68.56 & 89.47  &  38.39 \\
              & KNN \cite{sun2022out} & \textbf{76.18} & 71.10 & 70.87 & 90.18 & 34.13 \\            
              & ViM \cite{djurisic2022extremely} & \textbf{76.18} & 72.08 & 71.35 & 92.68 & \textbf{24.67} \\
              & RTL++ \cite{fan2024test} & \textbf{76.18} & 76.49 ± 0.49 & 71.49 ± 0.35& 90.53 ± 0.39& 44.39 ± 1.19 \\
              & TULIP \cite{anonymous2024TULIP} & \textbf{76.18} & \textbf{77.52 ± 0.06} & \textbf{64.96 ± 0.08} & 88.03 ± 0.02 & 48.01 ± 0.02 \\
              \rowcolor{gray!25}&  OODD & \textbf{76.18} & 72.66 ± 0.16 & 71.41 ± 0.27  & \textbf{93.84 ± 0.44} & 27.63 ± 1.16 \\
               \bottomrule
        \end{tabular}%
        }
        
      \label{tab:main_result-imagenet}%
    \end{table*}%

\subsection{Experiments Setup} \label{setup}
\textbf{Datasets.} We use four popular ID datasets CIFAR-10/100  \cite{krizhevsky2009learning}, ImageNet-200/1K \cite{deng2009imagenet}.
Following OpenOOD benchmark \cite{zhang2023openood}, the OOD testing datasets are categorized into two groups: Near OOD and Far OOD. Specifically, for CIFAR-10/100 \cite{krizhevsky2009learning} benchmarks, the Far OOD group includes MNIST \cite{lecun1998gradient}, SVHN \cite{netzer2011reading}, Textures \cite{cimpoi2014describing}, Places365 \cite{zhou2017places}, and the Near OOD group comprises CIFAR-100/10 and Tiny ImageNet-200 \cite{le2015tiny}. For ImageNet-200/1K, the Near OOD group includes SSB-hard \cite{vaze2021open}, NINCO \cite{bitterwolf2023or}, and the Far OOD group comprises iNaturalist \cite{van2018inaturalist}, Textures \cite{cimpoi2014describing}, and OpenImage-O \cite{wang2022vim}.

\noindent\textbf{Evaluation Metrics.} We use three metrics to evaluate the results: (1) FPR95, the false positive rate for OOD when the ID true positive rate is 95\%; (2) AUROC, the area under the receiver operating characteristic curve; and (3) ID ACC, the ID classification accuracy. Our reported results are averaged over three independent runs with different random seeds.

\noindent\textbf{Pre-trained Model.} For CIFAR-10/100 and ImageNet-200, we employed the ResNet-18 model \cite{he2016deep} trained with empirical risk minimization
 on each ID training set provided by OpenOOD \cite{zhang2023openood}. For ImageNet-1K, we use a pre-trained
ResNet-50 model \cite{he2016identity} from the PyTorch. 

\noindent\textbf{Baseline Methods.} 
We compared our method with various post-hoc methods, including MaxLogits \cite{hendrycks2019scaling}, ODIN \cite{liang2017enhancing}, Mahalanobis Distance (MD)\cite{lee2018simple}, Energy \cite{liu2020energy}, KNN \cite{sun2022out}, ViM \cite{djurisic2022extremely}, RTL++ \cite{fan2024test}, and TULIP \cite{anonymous2024TULIP}. In addition,
to the best of our knowledge, TULIP—the test-time method achieving state-of-the-art performance on the OpenOOD benchmark—was also included in our comparison.

\noindent\textbf{Remark on the implementation.} We implemented our method in PyTorch and conducted all experiments on a single NVIDIA GeForce RTX 4090 GPU. 
The ID dictionary only needs to be extracted once before deploying the trained model. Additionally, unlike KNN-based OOD scoring, which uses faiss.IndexFlatL2 with Euclidean distance, we compute the OOD score through GPU-accelerated matrix multiplication in PyTorch to obtain cosine similarity scores directly, so our approach is more memory-efficient and faster than the
former.
In all experiments, we set the batch size to 512.
In the main results presented in Tables \ref{tab:main_result-cifar} and \ref{tab:main_result-imagenet}, we use an inlier sampling ratio of $\alpha\% = 50\%$ with $M=4$, $\mathbb{\hat{K}} = 5$. For CIFAR-10/100, due to the smaller dataset size, we set the priority queue size $l$ to 128/512 respectively, and the memory bank size $mb$ is 5. Additionally, for CIFAR-10, we use $\mathbb{K} = 5$ and for CIFAR-100, $\mathbb{K}=10$. For larger datasets ImageNet-200/1K, we increase $l$ to 2048 and $mb$ to 128, and $\mathbb{K}=100$.

\section{Main Results.} 
\noindent\textbf{CIFAR Benchmarks.}
Firstly, we compare the proposed methods using C-Out (Cropping Outliers) with state-of-the-art OOD detection methods. The main experimental results on CIFAR-10/100 are shown in Table \ref{tab:main_result-cifar}, from which the following conclusions can be drawn:
(i) The proposed method OODD outperforms other methods in most cases. For example, on CIFAR-100, the Far OOD AUROC of our method reaches 93.64\%, while the second-best method (KNN) achieves 82.40\%.
(ii) Our method maintains a comparable ability for Near OOD detection. While the tail distributions of Near OOD and ID samples overlap considerably, our proposed approach does not significantly impair Near OOD detection performance. On CIFAR-100, our method even achieves state-of-the-art AUROC, though it does not lead in FPR95. However, AUROC provides a more comprehensive assessment than FPR95, which evaluates only a single point.

\noindent\textbf{ImageNet Benchmarks.}
For ImageNet-200/1K, prior works \cite{huang2021mos,wang2023out} show that many advanced methods developed on CIFAR benchmarks struggle with the large semantic space of ImageNet. Nevertheless, as shown in Table \ref{tab:main_result-imagenet}, our method achieves leading performance in both Near OOD and Far OOD detection on ImageNet-200. Additionally, although KNN performs moderately well on ImageNet-1K, our method significantly enhances performance in Far OOD without compromising its near OOD detection capability.

\subsection{Ablation Study}
In this section, we conduct ablation experiments on ImageNet-200 to analyze the effectiveness of each module and explore the impact of various hyperparameters. The main ablation study results are shown in Table \ref{table:ablation_main}, which demonstrate the effectiveness of our proposed informative inlier sampling and dynamic OOD dictionary strategies. 
\begin{table}[ht]
\small
\centering
\caption{The main ablation study results on ImageNet-200. IIS denotes informative inlier sampling, PR denotes priority queue, and MB denotes memory bank.}
\resizebox{.99\columnwidth}{!}{
\begin{tabular}{@{}ccc|cc|cc@{}}
\toprule
 \multirow{2}[1]{*}{\makecell{IIS}}   & \multirow{2}[1]{*}{\makecell{PR}} & \multirow{2}[1]{*}{\makecell{MB}}   & \multicolumn{2}{c|}{Near OOD} &\multicolumn{2}{c}{Far OOD} \\&&& FPR95 & AUROC & FPR95 & AUROC \\
  \midrule
  \xmark & \xmark & \xmark &   60.57   &  80.73 &  30.35  & 91.47 \\
\cmark & \xmark & \xmark &   58.92   &  81.67 & 26.97  &92.83 \\
\xmark & \cmark & \xmark &   53.85  & 85.69  & 21.47 & 95.45 \\
\xmark & \xmark & \cmark &   59.14   &   81.30 & 37.09 & 88.89\\
\cmark & \cmark & \cmark &   \textbf{53.70}   &  \textbf{85.74}  &  \textbf{20.89} &  \textbf{95.47} \\
 \bottomrule
\end{tabular}
}
\label{table:ablation_main}
\end{table}

\begin{figure}[ht]
  \centering
  \includegraphics[scale=0.33]{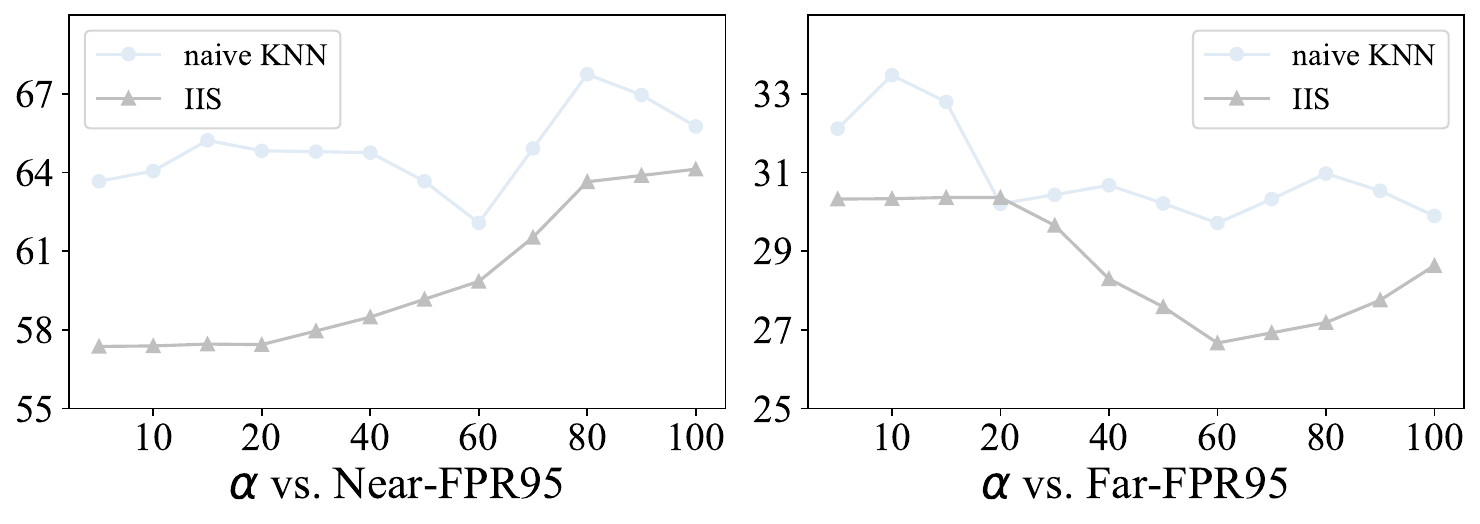}
  \vspace{-6mm}
  \caption{
    Results of varying $\alpha$ on ImageNet-200 ID. IIS refers to the implementation of Informative Inlier Sampling based on the naive KNN method \cite{sun2022out}. The value of $\alpha$ is on the horizontal axis, and the FPR95 is on the vertical axis.
  }
  \label{fig:FPR}
  \vspace{-2mm}
\end{figure}

\noindent\textbf{The effectiveness of informative inlier sampling.}
To examine this, we employ a naive KNN OOD detection \cite{sun2022out}, adjusting both the sampling ratio $\alpha\%$ and the hyper-parameter $\mathbb{K}$, which defines the $\mathbb{K}$-th largest cosine similarity in \( S_{in}(\bx^*) \). 
When choosing sampling ratio $\alpha\%$ for training
data for nearest neighbor search, $\mathbb{K}$ is selected from \{1, 5, 10, 20, 50, 100, 200, 500, 1000\} accordingly to the validation dataset OpenImage-O.
As shown in Figure \ref{fig:FPR}, the ablation results highlight how variations in $\alpha$ influence performance.

\begin{figure}[ht]
\centering
\centering
\includegraphics[width=0.4\textwidth]{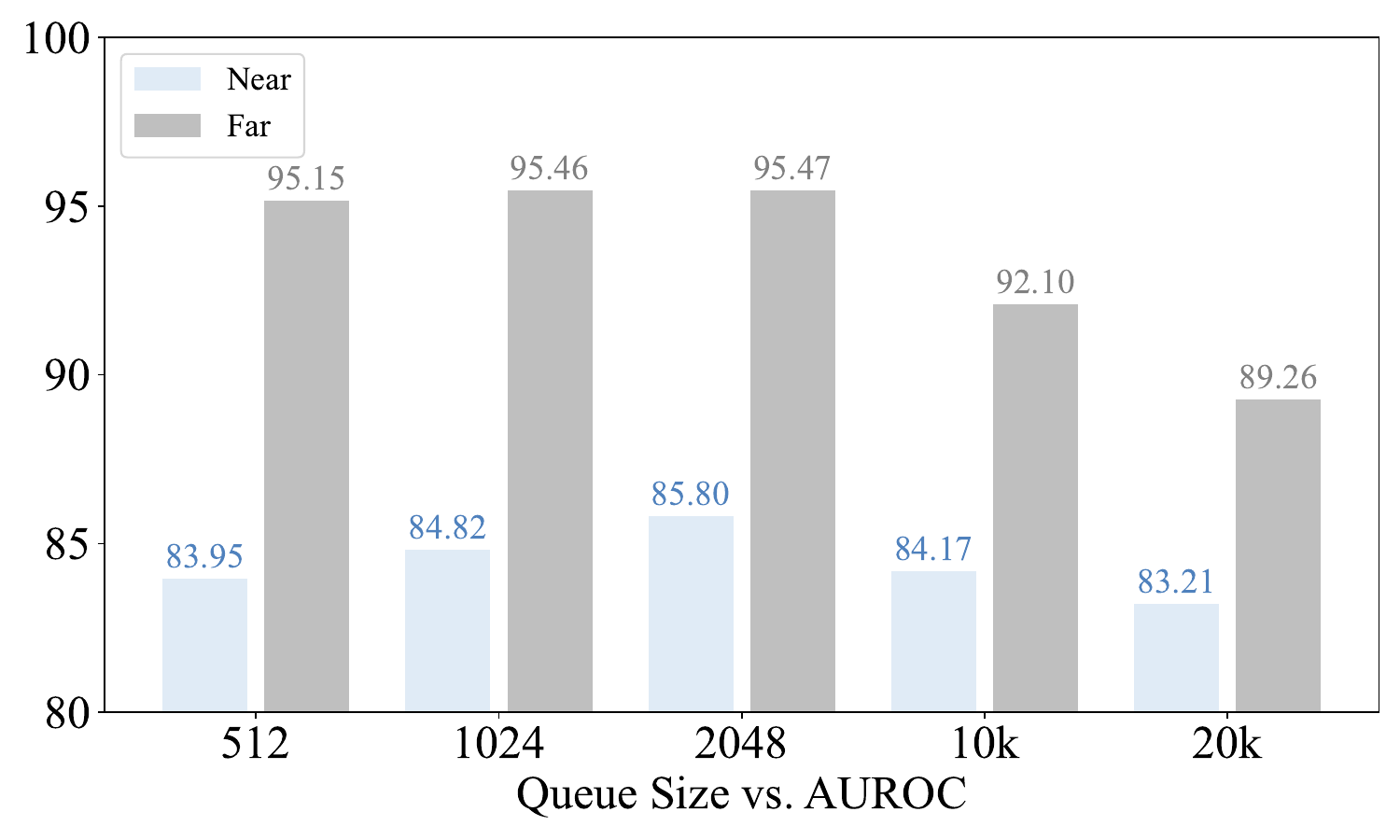}
\vspace{-0.4cm}
\caption{Comparison of AUROC detection performance using different queue sizes.}
\vspace{-0.7cm}
\label{fig:queue_size}
\end{figure}

\noindent\textbf{The effectiveness of the priority queue.}
In Figure \ref{fig:queue_size}, we systematically analyze the effect of the priority queue size $l$ under the conditions of not using a memory bank, with a batch size of 512. We vary the queue size $l$. Several key observations can be made: (i) In non-extreme cases, our method is not sensitive to $l$, with the results for both Near OOD and Far OOD showing a significantly lower FPR95 compared to the naive KNN method. (ii) When $l$ is particularly large, the effect of the priority queue becomes less prominent. Under this situation, the accumulation of ID data being added to the OOD dictionary increases, which leads to a decline in OOD detection.

\begin{figure}[ht]
\centering
\includegraphics[width=0.4\textwidth]{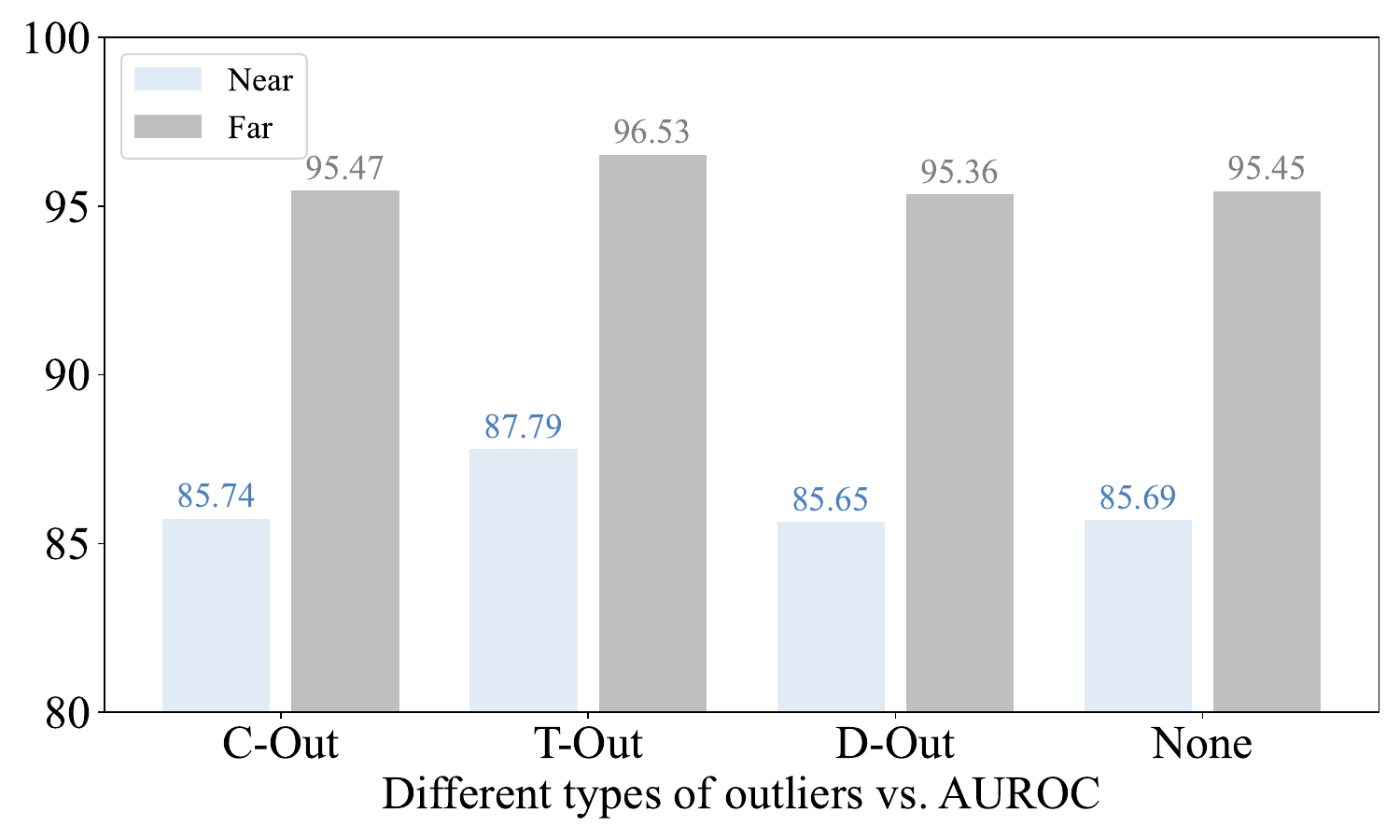}
\vspace{-0.4cm}
\caption{AUROC comparison using different types of outliers (C-Out, T-Out, D-Out, None) to initialize the OOD dictionary.}
\label{fig:outlier_ini}
\end{figure}

\begin{figure*}[ht]
\begin{center}
  \includegraphics[scale=0.43]{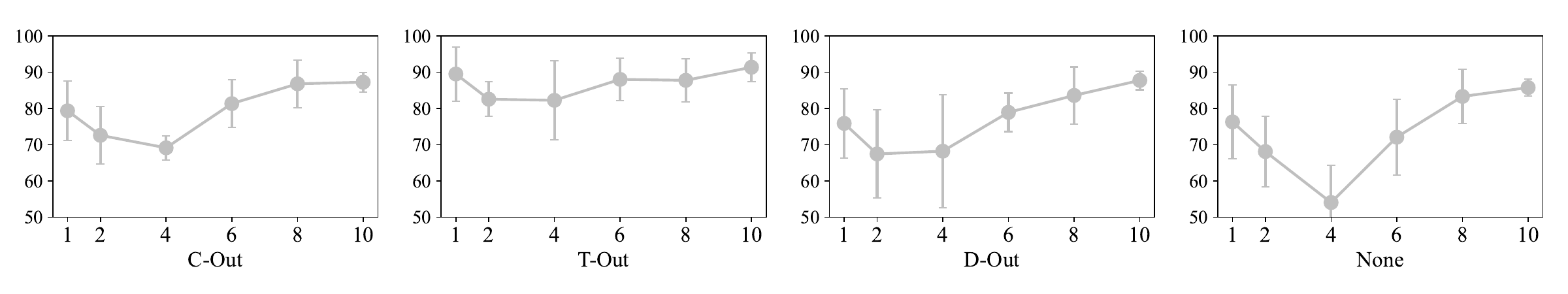}
  \vspace{-7mm}
\caption{
AUROC of test-time performance over the first 10 iterations for 10,000 ID samples from ImageNet-200 mixed with 100 OOD samples from OpenImage-O, averaged across five experimental runs with standard deviation.
}
\label{fig:iteration}

\end{center}
\end{figure*}

\noindent\textbf{The different initializations of OOD dictionary.}
In Figure \ref{fig:outlier_ini}, we show results for initializing the OOD dictionary on the ImageNet-200 benchmark with various types of outliers (C-Out, T-Out, D-Out, None). The experimental results indicate that the choice of initialization method has minimal impact on overall performance. To further investigate how initialization methods affect test-time performance, we set up a new experiment: selecting 10,000 samples from ImageNet-200 as ID dataset and mixing them with 100 OOD samples from OpenImage-O, shuffled randomly for testing. This experiment was repeated five times, and we recorded the AUROC for each of the first 10 iterations, as shown in Figure \ref{fig:iteration}. The results reveal that, without outliers to initialize the prior queue, OOD detection exhibits larger fluctuations in the first few iterations. However, as the iterations progress, AUROC values stabilize across different initialization methods, showing minimal differences. Notably, while the lack of prior OOD knowledge results in lower early performance compared to T-Out, our proposed C-Out initialization method achieves the most stable performance.
More experimental results can be seen in Table \ref{tab:breakdown} of the supplementary material.

\noindent\textbf{How to choose $\mathbb{K}$ and $\mathbb{\hat{K}}$?}
Following prior work \cite{sun2022out} and the OpenOOD benchmark \cite{zhang2023openood}, we first select the parameter $\mathbb{K}$ based on the OOD validation set provided by OpenOOD. Specifically, for CIFAR-10/100, the OOD validation set is tiny ImageNet-200, while for ImageNet-200/1K, the OOD validation set is OpenImage-O. All validation set data does not overlap with the test set data. Then, we are surprised to find that a relatively small value of $\mathbb{\hat{K}}$ yielded strong performance across all experiments. As a result, we fixed $\mathbb{\hat{K}}=5$ for all our experiments.
\begin{table}[!ht]
    \centering
    \caption{Time cost comparison of different OOD detection methods.}
    \resizebox{\linewidth}{!}{
    \begin{tabular}{cccc}
    \toprule
        Method & CIFAR-10 & ImageNet-200 & Time cost factor \\
    \midrule
        MaxLogits &1.20 s&1.45 s& 1x\\
        KNN &9.61&30.17 s& 15x\\
        RTL++ & 82.07 s & 103.80 s& 70x\\
        OODD &3.98 s&10.13 s&5x\\
    \bottomrule
    \end{tabular}
    }
    
    \label{tab:cost_time}
\end{table}

\noindent\textbf{Time Cost.}
To provide a more intuitive and fair comparison of computational overhead in OOD detection, we specifically measured the time cost from obtaining the image encoder's features to generating the final OOD scores. As shown in Table \ref{tab:cost_time}, Maxlogits exhibits the shortest processing time as it only involves a single linear classifier operation. RTL++ incurs the longest time cost due to the need to solve a Lasso regression between the latent features and the OOD scores \cite{fan2024test}. Meanwhile, KNN demands 15 times more computational time than Maxlogits. In comparison to KNN, our proposed method reduces the time cost by a factor of 3, highlighting its practical applicability.

\section{Further Discussion}
\subsection{Integrate with the CLIP-based methods} 
When using the Contrastive Language-Image Pre-training (CLIP) model as the encoder \cite{radford2021learning}, our method can be effectively integrated into various CLIP-based OOD detection methods, such as MCM \cite{ming2022delving} and NegLabel \cite{jiang2024negative}. As shown in Table \ref{tab:clip}, when combined with these CLIP-based methods, OODD significantly enhances performance across most datasets.
\begin{table}[htbp]
  \centering

  \caption{OOD detection performance for ImageNet-1K as ID}
  
  \resizebox{1.0\linewidth}{!}{
  \begin{tabular}{ccccccccc}
   \toprule
                                             & \multicolumn{2}{c}{iNaturalist}           & \multicolumn{2}{c}{SUN}           & \multicolumn{2}{c}{Places} & \multicolumn{2}{c}{Texture}          \\
   Method   & FPR95$\downarrow$ & AUROC$\uparrow$ & FPR95$\downarrow$ & AUROC$\uparrow$ & FPR95$\downarrow$ & AUROC$\uparrow$ & FPR95$\downarrow$ &AUROC$\uparrow$\\  \toprule
   MCM      &    30.91  & 94.61 & 37.59 & 92.57 & 44.69  & 89.77  & 57.77  & 86.11 \\
   +OODD       &  2.22   &  99.36  &  21.49   &  95.01   &  44.76   &  87.10   &  30.69  &  93.27  \\
    Improve & \textcolor{lightgreen}{$\nabla$-28.69} & \textcolor{lightgreen}{$\triangle$+4.75} & \textcolor{lightgreen}{$\nabla$-16.10} & \textcolor{lightgreen}{$\triangle$+2.44} & \textcolor{orange}{$\triangle$+0.07} & \textcolor{orange}{$\nabla$-2.67} & \textcolor{lightgreen}{$\nabla$-27.08} & \textcolor{lightgreen}{$\triangle$+7.16} \\
     \hline
    NegLabel   & 1.91 &    99.49 & 20.53&95.49  &35.59&91.64 & 43.56 &90.22  \\
   +OODD       & 0.85    &  99.79  &  12.94   & 97.17    & 30.68    &  92.51   & 30.67   & 94.51  \\
    Improve & \textcolor{lightgreen}{$\nabla$-1.06} & \textcolor{lightgreen}{$\triangle$+0.30} & \textcolor{lightgreen}{$\nabla$-7.59} & \textcolor{lightgreen}{$\triangle$+1.68} & \textcolor{lightgreen}{$\nabla$-4.91} & \textcolor{lightgreen}{$\triangle$+0.87} & \textcolor{lightgreen}{$\nabla$-12.89} & \textcolor{lightgreen}{$\triangle$+4.29} \\
  \bottomrule
  \end{tabular}
  }
    \label{tab:clip}
   
  \end{table}

\subsection{The Temporal Drift OOD Scenarios} 
Since OODD is a test-time method, it is crucial to address temporal drift where OOD scenarios evolve over time. For example, consider a temporal drift with time points $t_0$, $t_1$, $t_2$, and $t_3$. The temporal drift of OOD scenarios goes from Textures ($t_0$-$t_1$) to Places365 ($t_1$-$t_2$) and then to SVHN ($t_2$-$t_3$), denoted as $T \rightarrow P \rightarrow S$ in Table \ref{tab:new_setting}. After incorporating our OODD method under the temporal drift OOD scenarios, we observe improvements across all datasets. Furthermore, the earlier a dataset is encountered, the greater the gain, demonstrating the effectiveness of our approach.

\begin{table}[htbp]
  \centering

  \caption{\small{Results for the temporal drift OOD scenario settings using CIFAR-100 as ID dataset. $T$ represents Texture, $P$ for Places365, and $S$ for SVHN.}}

  \resizebox{1.0\linewidth}{!}{
  \begin{tabular}{ccccccccc}
   \toprule
                                             & \multicolumn{2}{c}{Textures}           & \multicolumn{2}{c}{Places365}           & \multicolumn{2}{c}{SVHN} & \multicolumn{2}{c}{Mean}          \\
   Temporal Drift   & FPR95$\downarrow$ & AUROC$\uparrow$ & FPR95$\downarrow$ & AUROC$\uparrow$ & FPR95$\downarrow$ & AUROC$\uparrow$ & FPR95$\downarrow$ &AUROC$\uparrow$\\  \toprule
   $T \rightarrow P \rightarrow S$     &  \textbf{35.33}   &  \textbf{91.80}  & 53.61    &  83.27   &   18.68  &  97.13   &  \textbf{35.87}  & \textbf{90.73}  \\
   $P \rightarrow S \rightarrow T$     &   45.42  &  86.25  &  \textbf{50.60}    &  \textbf{85.16}   &  17.76   & 97.31   &  37.93 &  89.57  \\
    $S \rightarrow T \rightarrow P$  &   42.40  &  87.77  &  58.63   &   80.19  &  \textbf{8.12}   &  \textbf{98.69}   & 36.39   & 88.88  \\
  \hline

    w/o OODD &  53.56   & 83.66   &   60.70  &  79.43   &   51.75  &   84.15  &  55.34  & 82.41  \\
   \bottomrule
  \end{tabular}
  }
    \label{tab:new_setting}
   
  \end{table}
  
\subsection{Combine with More Post-hoc Methods}
\textbf{Our approach further enhances the effectiveness of various post-hoc OOD detection methods.} We explore multiple post-hoc methods to identify latent OOD features. As shown in Table \ref{tab:others}, models equipped with the OOD dictionary consistently outperform their counterparts (i.e., original OOD detection performance), validating the effectiveness of the OOD dictionary mechanism. Notably, KNN with the OOD dictionary demonstrates a significant improvement, while the gains for MaxLogits and Energy methods are relatively modest. This discrepancy may be attributed to our OOD dictionary storing latent features, which have limited benefits for logits-based OOD detection methods. We hope that future work will focus on designing a more comprehensive OOD dictionary to enhance these methods further.

\begin{table}[ht]
\centering
\caption{AUROC performance comparison for whether the model is equipped with our proposed OOD dictionary.}
 \resizebox{1.0\linewidth}{!}{
\begin{tabular}{lcccccccc}
\toprule
In Dataset & \multicolumn{2}{c}{CIFAR-10} & \multicolumn{2}{c}{CIFAR-100} & \multicolumn{2}{c}{ImageNet-200} & \multicolumn{2}{c}{ImageNet-1K} \\
Method & Near & Far & Near & Far & Near & Far & Near & Far\\
\hline\hline
MaxLogits & 87.52 & 91.10 & 81.05 & 79.67 & 82.90 & 91.11& 76.46& 89.57\\
+Ours & 88.34 & 91.36 & 81.01 &86.56 &83.87 &92.03  &77.44 & 90.08\\
Improve & \textcolor{lightgreen}{$\triangle$+0.82} & \textcolor{lightgreen}{$\triangle$+0.26} & \textcolor{orange}{$\nabla$-0.04} & \textcolor{lightgreen}{$\triangle$+6.89} & \textcolor{lightgreen}{$\triangle$+0.97} & \textcolor{lightgreen}{$\triangle$+0.92} & \textcolor{lightgreen}{$\triangle$+0.98} & \textcolor{lightgreen}{$\triangle$+0.51}\\
\hline
Energy & 87.58 &91.21  & 80.91 &79.77 & 82.50& 90.86&  75.89 & 89.47 \\
+Ours & 88.41 & 91.59 & 80.98 & 87.74 &83.61 &91.86 & 76.92 & 89.97\\
Improve & \textcolor{lightgreen}{$\triangle$+0.83} & \textcolor{lightgreen}{$\triangle$+0.38} & \textcolor{lightgreen}{$\triangle$+0.07} & \textcolor{lightgreen}{$\triangle$+7.97} & \textcolor{lightgreen}{$\triangle$+1.11} & \textcolor{lightgreen}{$\triangle$+1.00} & \textcolor{lightgreen}{$\triangle$+1.03} & \textcolor{lightgreen}{$\triangle$+0.50} \\
\hline
KNN & 90.64 & 92.96 & 80.18 &82.40 &81.57 &93.16 &71.10 & 90.18\\
+Ours & 90.96 & 95.77 & 82.10 &93.64 &85.74 & 95.47& 72.66& 93.84\\
Improve & \textcolor{lightgreen}{$\triangle$+0.32} & \textcolor{lightgreen}{$\triangle$+2.81} & \textcolor{lightgreen}{$\triangle$+1.92} & \textcolor{lightgreen}{$\triangle$+11.24} & \textcolor{lightgreen}{$\triangle$+4.17} & \textcolor{lightgreen}{$\triangle$+2.31} & \textcolor{lightgreen}{$\triangle$+1.56} & \textcolor{lightgreen}{$\triangle$+3.66} \\
\bottomrule
\end{tabular}
}
\label{tab:others}
\end{table}

\subsection{A Closer Look at KNN-based OOD
Detection}
\textbf{For OOD detection, cosine similarity equals Euclidean distance after feature normalization but is more efficient.}
The KNN-based OOD detection \cite{sun2022out} shows that ID data tends to have larger Euclidean distances before normalization, which undermines assumptions about KNN distances in high-dimensional space. Fortunately, the latent features of ID data have larger $\ell_2$-norm values than those of OOD data, which effectively mitigates the dimension curse when calculating Euclidean distances.
Thus, given two vectors $\mathbf{z}_1$ and $\mathbf{z}_2$, we first apply $\ell_2$-norm normalization to each vector to obtain their normalized forms, $\hat{\mathbf{z}}_1$ and $\hat{\mathbf{z}}_2$, where $\hat{\mathbf{z}}_i=\frac{\mathbf{z}_i}{\left\|\mathbf{z}_i\right\|_2}$ for $i=1,2$. After normalization, the Euclidean distance and cosine similarity between $\hat{\mathbf{z}}_1$ and $\hat{\mathbf{z}}_2$ are closely related. For the above two normalized vectors, the Euclidean distance $d_E$ can be expressed as a function of their cosine similarity as follows:
\begin{equation}
    d_E\left(\hat{\mathbf{z}}_1, \hat{\mathbf{z}}_2\right)=\left\|\hat{\mathbf{z}}_1-\hat{\mathbf{z}}_2\right\|_2=\sqrt{2-2 \cos \left(\mathbf{z}_1, \mathbf{z}_2\right)}
\end{equation}
We are surprised to find that the cosine similarity $\cos \left(\mathbf{z}_1, \mathbf{z}_2\right)$ used in our method has a monotonic relationship with OOD score $-d_E\left(\hat{\mathbf{z}}_1, \hat{\mathbf{z}}_2\right)$ employed by KNN. For OOD detection, this monotonic relationship implies that the two functions are equivalent at the decision level. Moreover, after normalization, directly using $\cos \left(\mathbf{z}_1, \mathbf{z}_2\right)$ only requires dot product operations, significantly reducing computational overhead, especially when calculated in parallel.

\begin{figure}[ht]
  \centering
  \includegraphics[scale=0.22]{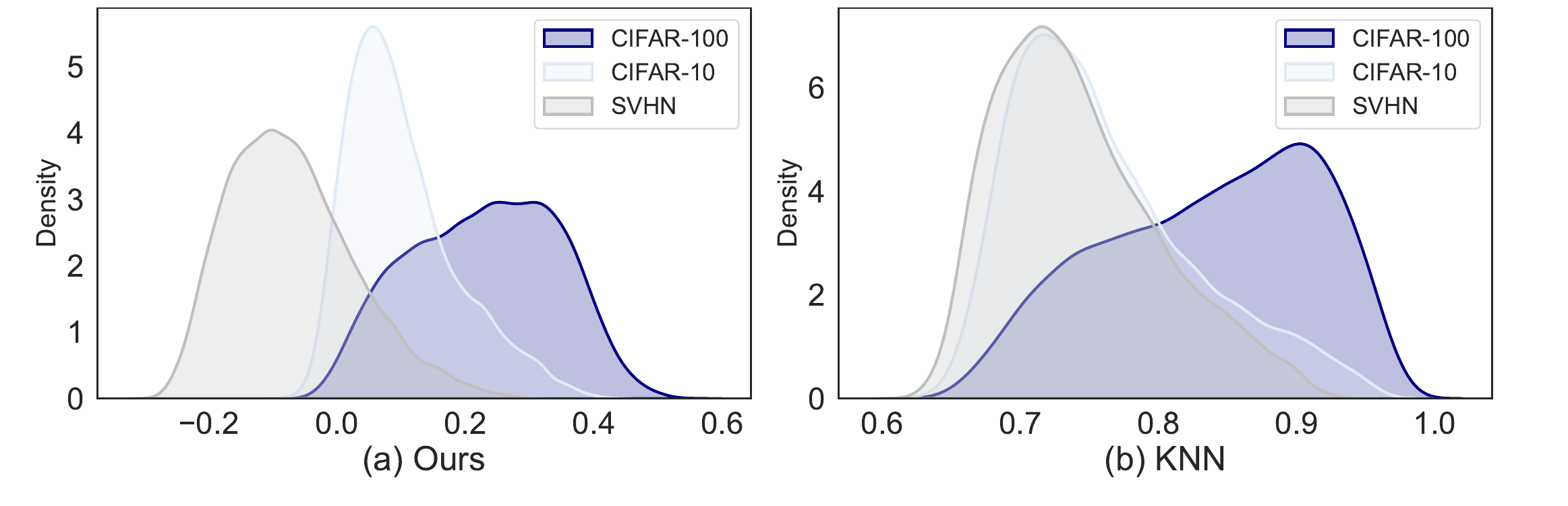}
  \vspace{-6mm}
  \caption{
    The density of the obtained ID and OOD score with the
proposed method (left) and KNN \cite{sun2022out} (right). CIFAR-100 is ID, Near OOD is CIFAR-10, and SVHN is Far OOD.
  }
  \label{fig:kde}
  \vspace{-4mm}
\end{figure}

\noindent\textbf{OOD Score Density: Our Method vs. KNN.}
We further use CIFAR-100 as ID and plot the density of the OOD scores using a kernel density estimate (KDE) to analyze the impact of Near OOD (CIFAR-10) and Far OOD (SVHN). Besides, we provide a theoretical analysis comparing our method with KNN as an OOD score function for OOD detection in Section \ref{theory_analysis} of the Supplementary Material. We observe that the proposed method effectively captures the features of OOD samples. Specifically, in Figure \ref{fig:kde}, we make the following observations:
(i) Thanks to the calibration of OOD scores by the OOD dictionary, the distribution of ID scores is more concentrated.
(ii) In Figure \ref{fig:kde}(b), we observe that for the left-tail distribution, the intersection over union (IoU) between Far OOD and ID is smaller. As a result, the overall distribution of Far OOD scores shifts more to the left, achieving greater calibration gains.

\section{Conclusion}
We present a test-time OOD detection method called OODD that leverages a dynamic dictionary of OOD features to achieve high detection performance without fine-tuning or complex calibration processes. By integrating a priority queue and employing inlier and outlier sampling strategies, OODD effectively adapts to diverse OOD distributions. Our method not only outperforms the state-of-the-art OOD detection method but also demonstrates strong computational efficiency, as verified on the OpenOOD benchmark. These findings highlight OODD’s potential to enhance model uncertainty awareness at test time—an essential capability for deploying robust models in open-world scenarios.

{
    \small
    \bibliographystyle{ieeenat_fullname}
    \bibliography{main}

\begin{thebibliography}{41}
\providecommand{\natexlab}[1]{#1}
\providecommand{\url}[1]{\texttt{#1}}
\expandafter\ifx\csname urlstyle\endcsname\relax
  \providecommand{\doi}[1]{doi: #1}\else
  \providecommand{\doi}{doi: \begingroup \urlstyle{rm}\Url}\fi

\bibitem[Anonymous(2024)]{anonymous2024TULIP}
Anonymous.
\newblock {TUL}ip: Test-time uncertainty estimation via linearization and weight perturbation.
\newblock In \emph{Submitted to The Thirteenth International Conference on Learning Representations}, 2024.
\newblock under review.

\bibitem[Bai et~al.(2023)Bai, Canal, Du, Kwon, Nowak, and Li]{bai2023feed}
Haoyue Bai, Gregory Canal, Xuefeng Du, Jeongyeol Kwon, Robert~D Nowak, and Yixuan Li.
\newblock Feed two birds with one scone: Exploiting wild data for both out-of-distribution generalization and detection.
\newblock In \emph{International Conference on Machine Learning}, pages 1454--1471. PMLR, 2023.

\bibitem[Bitterwolf et~al.(2023)Bitterwolf, M{\"u}ller, and Hein]{bitterwolf2023or}
Julian Bitterwolf, Maximilian M{\"u}ller, and Matthias Hein.
\newblock In or out? fixing imagenet out-of-distribution detection evaluation.
\newblock \emph{arXiv preprint arXiv:2306.00826}, 2023.

\bibitem[Chen et~al.(2021)Chen, Li, Wu, Liang, and Jha]{chen2021atom}
Jiefeng Chen, Yixuan Li, Xi Wu, Yingyu Liang, and Somesh Jha.
\newblock Atom: Robustifying out-of-distribution detection using outlier mining.
\newblock In \emph{Machine Learning and Knowledge Discovery in Databases. Research Track: European Conference, ECML PKDD 2021, Bilbao, Spain, September 13--17, 2021, Proceedings, Part III 21}, pages 430--445. Springer, 2021.

\bibitem[Cimpoi et~al.(2014)Cimpoi, Maji, Kokkinos, Mohamed, and Vedaldi]{cimpoi2014describing}
Mircea Cimpoi, Subhransu Maji, Iasonas Kokkinos, Sammy Mohamed, and Andrea Vedaldi.
\newblock Describing textures in the wild.
\newblock In \emph{Proceedings of the IEEE conference on computer vision and pattern recognition}, pages 3606--3613, 2014.

\bibitem[Deng et~al.(2009)Deng, Dong, Socher, Li, Li, and Fei-Fei]{deng2009imagenet}
Jia Deng, Wei Dong, Richard Socher, Li-Jia Li, Kai Li, and Li Fei-Fei.
\newblock Imagenet: A large-scale hierarchical image database.
\newblock In \emph{2009 IEEE conference on computer vision and pattern recognition}, pages 248--255. Ieee, 2009.

\bibitem[Djurisic et~al.(2022)Djurisic, Bozanic, Ashok, and Liu]{djurisic2022extremely}
Andrija Djurisic, Nebojsa Bozanic, Arjun Ashok, and Rosanne Liu.
\newblock Extremely simple activation shaping for out-of-distribution detection.
\newblock \emph{arXiv preprint arXiv:2209.09858}, 2022.

\bibitem[Fan et~al.(2024)Fan, Liu, Qiu, Wang, Huai, Shangguan, Gou, Liu, Fu, Fu, et~al.]{fan2024test}
Ke Fan, Tong Liu, Xingyu Qiu, Yikai Wang, Lian Huai, Zeyu Shangguan, Shuang Gou, Fengjian Liu, Yuqian Fu, Yanwei Fu, et~al.
\newblock Test-time linear out-of-distribution detection.
\newblock In \emph{Proceedings of the IEEE/CVF Conference on Computer Vision and Pattern Recognition}, pages 23752--23761, 2024.

\bibitem[Gonz{\'a}lez et~al.(2022)Gonz{\'a}lez, Gotkowski, Fuchs, Bucher, Dadras, Fischbach, Kaltenborn, and Mukhopadhyay]{gonzalez2022distance}
Camila Gonz{\'a}lez, Karol Gotkowski, Moritz Fuchs, Andreas Bucher, Armin Dadras, Ricarda Fischbach, Isabel~Jasmin Kaltenborn, and Anirban Mukhopadhyay.
\newblock Distance-based detection of out-of-distribution silent failures for covid-19 lung lesion segmentation.
\newblock \emph{Medical image analysis}, 82:\penalty0 102596, 2022.

\bibitem[Goodfellow et~al.(2014)Goodfellow, Shlens, and Szegedy]{goodfellow2014explaining}
Ian~J Goodfellow, Jonathon Shlens, and Christian Szegedy.
\newblock Explaining and harnessing adversarial examples.
\newblock \emph{arXiv preprint arXiv:1412.6572}, 2014.

\bibitem[He et~al.(2016{\natexlab{a}})He, Zhang, Ren, and Sun]{he2016deep}
Kaiming He, Xiangyu Zhang, Shaoqing Ren, and Jian Sun.
\newblock Deep residual learning for image recognition.
\newblock In \emph{Proceedings of the IEEE conference on computer vision and pattern recognition}, pages 770--778, 2016{\natexlab{a}}.

\bibitem[He et~al.(2016{\natexlab{b}})He, Zhang, Ren, and Sun]{he2016identity}
Kaiming He, Xiangyu Zhang, Shaoqing Ren, and Jian Sun.
\newblock Identity mappings in deep residual networks.
\newblock In \emph{Computer Vision--ECCV 2016: 14th European Conference, Amsterdam, The Netherlands, October 11--14, 2016, Proceedings, Part IV 14}, pages 630--645. Springer, 2016{\natexlab{b}}.

\bibitem[Hendrycks and Gimpel(2016)]{hendrycks2016baseline}
Dan Hendrycks and Kevin Gimpel.
\newblock A baseline for detecting misclassified and out-of-distribution examples in neural networks.
\newblock \emph{arXiv preprint arXiv:1610.02136}, 2016.

\bibitem[Hendrycks et~al.(2018)Hendrycks, Mazeika, and Dietterich]{hendrycks2018deep}
Dan Hendrycks, Mantas Mazeika, and Thomas Dietterich.
\newblock Deep anomaly detection with outlier exposure.
\newblock \emph{arXiv preprint arXiv:1812.04606}, 2018.

\bibitem[Hendrycks et~al.(2019)Hendrycks, Basart, Mazeika, Zou, Kwon, Mostajabi, Steinhardt, and Song]{hendrycks2019scaling}
Dan Hendrycks, Steven Basart, Mantas Mazeika, Andy Zou, Joe Kwon, Mohammadreza Mostajabi, Jacob Steinhardt, and Dawn Song.
\newblock Scaling out-of-distribution detection for real-world settings.
\newblock \emph{arXiv preprint arXiv:1911.11132}, 2019.

\bibitem[Huang and Li(2021)]{huang2021mos}
Rui Huang and Yixuan Li.
\newblock Mos: Towards scaling out-of-distribution detection for large semantic space.
\newblock In \emph{Proceedings of the IEEE/CVF Conference on Computer Vision and Pattern Recognition}, pages 8710--8719, 2021.

\bibitem[Jiang et~al.(2024)Jiang, Liu, Fang, Chen, Liu, Zheng, and Han]{jiang2024negative}
Xue Jiang, Feng Liu, Zhen Fang, Hong Chen, Tongliang Liu, Feng Zheng, and Bo Han.
\newblock Negative label guided ood detection with pretrained vision-language models.
\newblock \emph{arXiv preprint arXiv:2403.20078}, 2024.

\bibitem[Krizhevsky et~al.(2009)Krizhevsky, Hinton, et~al.]{krizhevsky2009learning}
Alex Krizhevsky, Geoffrey Hinton, et~al.
\newblock Learning multiple layers of features from tiny images.
\newblock 2009.

\bibitem[Le and Yang(2015)]{le2015tiny}
Ya Le and Xuan Yang.
\newblock Tiny imagenet visual recognition challenge.
\newblock \emph{CS 231N}, 7\penalty0 (7):\penalty0 3, 2015.

\bibitem[LeCun et~al.(1998)LeCun, Bottou, Bengio, and Haffner]{lecun1998gradient}
Yann LeCun, L{\'e}on Bottou, Yoshua Bengio, and Patrick Haffner.
\newblock Gradient-based learning applied to document recognition.
\newblock \emph{Proceedings of the IEEE}, 86\penalty0 (11):\penalty0 2278--2324, 1998.

\bibitem[Lee et~al.(2018)Lee, Lee, Lee, and Shin]{lee2018simple}
Kimin Lee, Kibok Lee, Honglak Lee, and Jinwoo Shin.
\newblock A simple unified framework for detecting out-of-distribution samples and adversarial attacks.
\newblock \emph{Advances in neural information processing systems}, 31, 2018.

\bibitem[Liang et~al.(2017)Liang, Li, and Srikant]{liang2017enhancing}
Shiyu Liang, Yixuan Li, and Rayadurgam Srikant.
\newblock Enhancing the reliability of out-of-distribution image detection in neural networks.
\newblock \emph{arXiv preprint arXiv:1706.02690}, 2017.

\bibitem[Liu et~al.(2020)Liu, Wang, Owens, and Li]{liu2020energy}
Weitang Liu, Xiaoyun Wang, John Owens, and Yixuan Li.
\newblock Energy-based out-of-distribution detection.
\newblock \emph{Advances in neural information processing systems}, 33:\penalty0 21464--21475, 2020.

\bibitem[Ming et~al.(2022)Ming, Cai, Gu, Sun, Li, and Li]{ming2022delving}
Yifei Ming, Ziyang Cai, Jiuxiang Gu, Yiyou Sun, Wei Li, and Yixuan Li.
\newblock Delving into out-of-distribution detection with vision-language representations.
\newblock \emph{Advances in neural information processing systems}, 35:\penalty0 35087--35102, 2022.

\bibitem[Netzer et~al.(2011)Netzer, Wang, Coates, Bissacco, Wu, and Ng]{netzer2011reading}
Yuval Netzer, Tao Wang, Adam Coates, Alessandro Bissacco, Bo Wu, and Andrew~Y Ng.
\newblock Reading digits in natural images with unsupervised feature learning.
\newblock 2011.

\bibitem[Nguyen et~al.(2015)Nguyen, Yosinski, and Clune]{nguyen2015deep}
Anh Nguyen, Jason Yosinski, and Jeff Clune.
\newblock Deep neural networks are easily fooled: High confidence predictions for unrecognizable images.
\newblock In \emph{Proceedings of the IEEE conference on computer vision and pattern recognition}, pages 427--436, 2015.

\bibitem[Radford et~al.(2021)Radford, Kim, Hallacy, Ramesh, Goh, Agarwal, Sastry, Askell, Mishkin, Clark, et~al.]{radford2021learning}
Alec Radford, Jong~Wook Kim, Chris Hallacy, Aditya Ramesh, Gabriel Goh, Sandhini Agarwal, Girish Sastry, Amanda Askell, Pamela Mishkin, Jack Clark, et~al.
\newblock Learning transferable visual models from natural language supervision.
\newblock In \emph{International conference on machine learning}, pages 8748--8763. PmLR, 2021.

\bibitem[Schmidt et~al.(2024)Schmidt, Schenk, Schwinn, and G{\"u}nnemann]{schmidt2024unified}
Sebastian Schmidt, Leonard Schenk, Leo Schwinn, and Stephan G{\"u}nnemann.
\newblock A unified approach towards active learning and out-of-distribution detection.
\newblock \emph{arXiv preprint arXiv:2405.11337}, 2024.

\bibitem[Shu et~al.(2022)Shu, Nie, Huang, Yu, Goldstein, Anandkumar, and Xiao]{shu2022test}
Manli Shu, Weili Nie, De-An Huang, Zhiding Yu, Tom Goldstein, Anima Anandkumar, and Chaowei Xiao.
\newblock Test-time prompt tuning for zero-shot generalization in vision-language models.
\newblock \emph{Advances in Neural Information Processing Systems}, 35:\penalty0 14274--14289, 2022.

\bibitem[Sun et~al.(2024)Sun, Shi, Yu, and Lin]{sun2024diversity}
Peng Sun, Bei Shi, Daiwei Yu, and Tao Lin.
\newblock On the diversity and realism of distilled dataset: An efficient dataset distillation paradigm.
\newblock In \emph{Proceedings of the IEEE/CVF Conference on Computer Vision and Pattern Recognition}, pages 9390--9399, 2024.

\bibitem[Sun et~al.(2022)Sun, Ming, Zhu, and Li]{sun2022out}
Yiyou Sun, Yifei Ming, Xiaojin Zhu, and Yixuan Li.
\newblock Out-of-distribution detection with deep nearest neighbors.
\newblock In \emph{International Conference on Machine Learning}, pages 20827--20840. PMLR, 2022.

\bibitem[Van~Horn et~al.(2018)Van~Horn, Mac~Aodha, Song, Cui, Sun, Shepard, Adam, Perona, and Belongie]{van2018inaturalist}
Grant Van~Horn, Oisin Mac~Aodha, Yang Song, Yin Cui, Chen Sun, Alex Shepard, Hartwig Adam, Pietro Perona, and Serge Belongie.
\newblock The inaturalist species classification and detection dataset.
\newblock In \emph{Proceedings of the IEEE conference on computer vision and pattern recognition}, pages 8769--8778, 2018.

\bibitem[Vaze et~al.(2021)Vaze, Han, Vedaldi, and Zisserman]{vaze2021open}
Sagar Vaze, Kai Han, Andrea Vedaldi, and Andrew Zisserman.
\newblock Open-set recognition: A good closed-set classifier is all you need?
\newblock \emph{arXiv preprint arXiv:2110.06207}, 2021.

\bibitem[Wang et~al.(2022)Wang, Li, Feng, and Zhang]{wang2022vim}
Haoqi Wang, Zhizhong Li, Litong Feng, and Wayne Zhang.
\newblock Vim: Out-of-distribution with virtual-logit matching.
\newblock In \emph{Proceedings of the IEEE/CVF conference on computer vision and pattern recognition}, pages 4921--4930, 2022.

\bibitem[Wang et~al.(2023)Wang, Ye, Liu, Dai, Kalander, Liu, Hao, and Han]{wang2023out}
Qizhou Wang, Junjie Ye, Feng Liu, Quanyu Dai, Marcus Kalander, Tongliang Liu, Jianye Hao, and Bo Han.
\newblock Out-of-distribution detection with implicit outlier transformation.
\newblock \emph{arXiv preprint arXiv:2303.05033}, 2023.

\bibitem[Yang et~al.(2023)Yang, Liang, Cao, and He]{yang2023auto}
Puning Yang, Jian Liang, Jie Cao, and Ran He.
\newblock Auto: Adaptive outlier optimization for online test-time ood detection.
\newblock \emph{arXiv preprint arXiv:2303.12267}, 2023.

\bibitem[Zhang et~al.(2023{\natexlab{a}})Zhang, Inkawhich, Linderman, Chen, and Li]{zhang2023mixture}
Jingyang Zhang, Nathan Inkawhich, Randolph Linderman, Yiran Chen, and Hai Li.
\newblock Mixture outlier exposure: Towards out-of-distribution detection in fine-grained environments.
\newblock In \emph{Proceedings of the IEEE/CVF Winter Conference on Applications of Computer Vision}, pages 5531--5540, 2023{\natexlab{a}}.

\bibitem[Zhang et~al.(2023{\natexlab{b}})Zhang, Yang, Wang, Wang, Lin, Zhang, Sun, Du, Zhou, Zhang, et~al.]{zhang2023openood}
Jingyang Zhang, Jingkang Yang, Pengyun Wang, Haoqi Wang, Yueqian Lin, Haoran Zhang, Yiyou Sun, Xuefeng Du, Kaiyang Zhou, Wayne Zhang, et~al.
\newblock Openood v1. 5: Enhanced benchmark for out-of-distribution detection.
\newblock \emph{arXiv preprint arXiv:2306.09301}, 2023{\natexlab{b}}.

\bibitem[Zhang et~al.(2024)Zhang, Feng, Zhou, Bian, Hu, and Zhang]{zhang2024best}
Qingyang Zhang, Qiuxuan Feng, Joey~Tianyi Zhou, Yatao Bian, Qinghua Hu, and Changqing Zhang.
\newblock The best of both worlds: On the dilemma of out-of-distribution detection.
\newblock \emph{arXiv preprint arXiv:2410.11576}, 2024.

\bibitem[Zhou et~al.(2017)Zhou, Lapedriza, Khosla, Oliva, and Torralba]{zhou2017places}
Bolei Zhou, Agata Lapedriza, Aditya Khosla, Aude Oliva, and Antonio Torralba.
\newblock Places: A 10 million image database for scene recognition.
\newblock \emph{IEEE transactions on pattern analysis and machine intelligence}, 40\penalty0 (6):\penalty0 1452--1464, 2017.

\bibitem[Zhu et~al.(2023)Zhu, Yu, Yao, Liu, Niu, Sugiyama, and Han]{zhu2023diversified}
Jianing Zhu, Geng Yu, Jiangchao Yao, Tongliang Liu, Gang Niu, Masashi Sugiyama, and Bo Han.
\newblock Diversified outlier exposure for out-of-distribution detection via informative extrapolation.
\newblock \emph{arXiv preprint arXiv:2310.13923}, 2023.

\end{thebibliography}
}
\clearpage
\appendix

\setcounter{page}{1}
\maketitlesupplementary
\appendix
\section{Acknowledgement}
This research is supported by National Natural Science Foundation of China under Grant (No.62106139) and State Key
Laboratory of High Performance Computing, National University of Defense Technology (No.202401-11). The authors also gratefully acknowledge the insightful comments provided by the anonymous reviewers.

\section{Experiments}
\subsection{More Detailed Results}
In Table \ref{tab:breakdown}, we provide
more detailed experimental results on the CIFAR and ImageNet benchmarks across various types of outliers (C-Out, T-Out, D-Out, None).
\subsection{Discussion of Computational Cost}
In Table \ref{tab:cost_time}, we present the computational overhead of our method. Unlike the naive KNN detection, our approach relies solely on dot products after normalization, significantly reducing overhead, especially with parallel computation. This is equivalent to adding a single linear layer, introducing an additional matrix multiplication with complexity $\mathcal{O}(d \cdot l)$ per sample, where $d$ is the feature dimension, and $l$ denotes the priority queue size. Updating the priority queue has a complexity of $\mathcal{O}(\log l)$. Since $l \leq 2048$ in all experiments, the extra overhead when integrating our method with others (e.g., energy-based or maxlogits) remains negligible.
\begin{table*}[!ht]
  \centering
  \caption{Detailed results of proposed approach performance with different outlier types across datasets. CIFAR-10/100 uses D-Out  from Tiny-ImageNet 597, whereas ImageNet-200 uses D-Out from ImageNet-800 \cite{zhang2023openood}.}
  \resizebox{0.80\linewidth}{!}{
  \begin{tabular}{cccccccccc}
   \toprule
                                                &               & \multicolumn{2}{c}{C-Out}           & \multicolumn{2}{c}{T-Out}           & \multicolumn{2}{c}{D-Out} & \multicolumn{2}{c}{None}          \\
  ID dataset                                    & OOD dataset   & FPR95$\downarrow$ & AUROC$\uparrow$ & FPR95$\downarrow$ & AUROC$\uparrow$ & FPR95$\downarrow$ & AUROC$\uparrow$ & FPR95$\downarrow$ & AUROC$\uparrow$\\  \toprule
  \multicolumn{1}{c}{\multirow{8}{*}{CIFAR-10}} & CIFAR-100     &40.66     &89.76    &44.70     &89.58     &47.24     &89.38     &43.14    &88.98   \\
  \multicolumn{1}{c}{}                          & Tiny-ImageNet &31.37     &92.16    &31.23     &92.56     &31.43     &92.58     &31.62    &91.89   \\
 \multicolumn{1}{c}{}                          &  \cellcolor{gray!25} Near OOD      &\cellcolor{gray!25}36.01     &\cellcolor{gray!25}90.96     &\cellcolor{gray!25}37.97     &\cellcolor{gray!25}91.07     &\cellcolor{gray!25}39.34     &\cellcolor{gray!25}90.98     &\cellcolor{gray!25}37.38    &\cellcolor{gray!25}90.44   \\
  \multicolumn{1}{c}{}                          & MNIST         &8.92     &98.01     &7.26      &98.35     &7.49     &98.24     &8.16     &98.15   \\
  \multicolumn{1}{c}{}                          & SVHN          &13.31     &96.73     &10.97     &97.35     &14.38     &96.57     &12.79    &96.92   \\
  \multicolumn{1}{c}{}                          & Textures      &20.16     &95.04     &18.61     &95.33     &21.14     &94.87     &23.64    &94.20   \\
  \multicolumn{1}{c}{}                          & Places365     &27.35     &93.32     &25.80     &93.92     &25.96     &93.90     &27.12    &93.55   \\
\multicolumn{1}{c}{} & \cellcolor{gray!25} Far OOD & \cellcolor{gray!25}17.44 & \cellcolor{gray!25}95.77 & \cellcolor{gray!25}15.66 & \cellcolor{gray!25}96.24 & \cellcolor{gray!25}17.24 & \cellcolor{gray!25}95.89 & \cellcolor{gray!25}17.93 & \cellcolor{gray!25}95.70 \\ \hline
  \multirow{8}{*}{CIFAR-100}                    & CIFAR-10      &58.60     &81.46     &59.12     &81.51     &61.22     &80.90     &61.09    &81.06   \\
                                                & Tiny-ImageNet &55.36     &82.73     &55.32     &82.78     &54.39     &82.96     &55.39    &82.74   \\
                                                 & \cellcolor{gray!25} Near OOD      &\cellcolor{gray!25}56.98     &\cellcolor{gray!25}82.10     &\cellcolor{gray!25}57.22     &\cellcolor{gray!25}82.15     &\cellcolor{gray!25}57.81     &\cellcolor{gray!25}81.93     &\cellcolor{gray!25}58.24    &\cellcolor{gray!25}81.90   \\
                                                & MNIST         &3.21     &99.37     &2.22      &99.53     &2.48     &99.48     &2.51    &99.48   \\
                                                & SVHN          &9.44     &98.54     &6.70     &98.90     &7.61     &98.78     &9.49    &98.55   \\
                                                & Textures      &35.75     &91.47     &34.37     &92.29     &34.84     &91.92     &34.40    &92.07   \\
                                                & Places365     &50.55     &85.14     &50.51     &85.35     &51.24     &85.09     &50.92    &85.20   \\
                                               & \cellcolor{gray!25} Far OOD       &\cellcolor{gray!25}24.74     &\cellcolor{gray!25}93.64     &\cellcolor{gray!25}23.45     &\cellcolor{gray!25}94.02     &\cellcolor{gray!25}24.04     &\cellcolor{gray!25}93.82     &\cellcolor{gray!25}24.33    &\cellcolor{gray!25}93.83   \\ \hline
  \multirow{7}{*}{ImageNet-200}                 & SSB-hard      &64.62     &81.76     &59.02     &84.41     &64.94     &81.54     &64.21    &81.66   \\
                                                & NINCO         &42.82     &89.72     &38.26     &91.17     &42.74     &89.76     &43.49    &89.71   \\
                                               & \cellcolor{gray!25} Near OOD      &\cellcolor{gray!25}53.70     &\cellcolor{gray!25}85.74     &\cellcolor{gray!25}48.64     &\cellcolor{gray!25}87.79     &\cellcolor{gray!25}53.84     &\cellcolor{gray!25}85.65     &\cellcolor{gray!25}53.85    &\cellcolor{gray!25}85.69   \\
                                                & iNaturalist   &9.14     &98.54     &3.32     &99.27      &10.48     &98.29     &9.73    &98.48   \\
                                                & Textures      &25.06     &95.24     &21.02     &96.21     &23.98     &95.46     &25.44    &95.23   \\
                                                & OpenImage-O   &28.44     &92.67     &26.13     &94.10     &29.04     &92.34     &29.22    &92.63   \\
                                                &  \cellcolor{gray!25}Far OOD       &\cellcolor{gray!25}20.89     &\cellcolor{gray!25}95.47     &\cellcolor{gray!25}16.83     &\cellcolor{gray!25}96.53     &\cellcolor{gray!25}21.17     &\cellcolor{gray!25}95.36     &\cellcolor{gray!25}21.47    &\cellcolor{gray!25}95.45   \\ \hline
  \multirow{7}{*}{ImageNet-1K}                  & SSB-hard      &80.35     &65.90     &77.59    &67.73     &   ~ - -~   &  ~ - -~    &80.36    &66.00   \\
                                                & NINCO         &62.48     &79.41     &59.80     &81.13     &  ~ - -~    &  ~ - -~    &63.06    &79.51   \\
                                               &  \cellcolor{gray!25}Near OOD      &\cellcolor{gray!25}71.41     &\cellcolor{gray!25}72.66     &\cellcolor{gray!25}68.70     &\cellcolor{gray!25}74.43     &\cellcolor{gray!25}   ~ - -~   &\cellcolor{gray!25}    ~ - -~  &\cellcolor{gray!25}71.71    &\cellcolor{gray!25}72.75   \\
                                                & iNaturalist   &21.54     &95.75     &13.14     &97.51     &   ~ - -~   &   ~ - -~   &24.30    &95.03   \\
                                                & Textures      &21.76     &96.18     &20.62     &96.49     &  ~ - -~    &   ~ - -~   &21.65    &96.25   \\
                                                & OpenIamge-O   &39.61     &89.60     &36.85     &91.25     &   ~ - -~   &   ~ - -~   &40.40    &89.12   \\
                                                & \cellcolor{gray!25} Far OOD       &\cellcolor{gray!25}27.63     &\cellcolor{gray!25}93.84     &\cellcolor{gray!25}23.54     &\cellcolor{gray!25}95.08     & \cellcolor{gray!25} ~ - -~    & \cellcolor{gray!25} ~ - -~    &\cellcolor{gray!25}28.79    &\cellcolor{gray!25}93.47   \\ \bottomrule
  \end{tabular}
  }
    \label{tab:breakdown}
  \end{table*}

\section{Theoretical Analysis} \label{theory_analysis}
\noindent\textbf{Setup.} The decision function for OOD detection using KNN \cite{sun2022out} is given by:

\begin{equation}
G\left(\mathbf{z}^* ; k\right)=\mathbf{1}\left\{-r_k\left(\mathbf{z}^*\right) \geq \lambda\right\},
\end{equation}
where $r_k\left(\mathbf{z}^*\right)=\left\|\mathbf{z}^*-\mathbf{z}_{(k)}\right\|_2$ represents the Euclidean distance to the k-th nearest neighbor, $\lambda$ is the threshold and $\mathbf{1}\{\cdot\}$ is the indicator function.
According to Bayes’ rule, the probability of $\mathbf{z_i}$ being ID data is:
\begin{equation}
 \begin{aligned}
\hat{p}\left(g_i=1 \mid \mathbf{z}_i\right) & =\frac{(1-\varepsilon) \hat{p}_{in }\left(\mathbf{z}_i\right)}{(1-\varepsilon) \hat{p}_{in }\left(\mathbf{z}_i\right)+\varepsilon \hat{p}_{out }\left(\mathbf{z}_i\right)} .
\end{aligned}
\end{equation}
where $p\left(g_i=0\right)=\varepsilon$, and the components are defined as: $\hat{p}_{i n}\left(\mathbf{z}_i \right)  =\frac{k}{c_b n\left(r_k\left(\mathbf{z}_i\right)\right)^{m-1}}$,where $c_b$ is constant, $\hat{p}_{\text {out }}\left(\mathbf{z}_i\right)=\hat{c}_0 \mathbf{1}\left\{\hat{p}_{\text {in }}\left(\mathbf{z}_i ;\right)<\frac{\beta \varepsilon \hat{c}_0}{(1-\beta)(1-\varepsilon)}\right\}$. If the OOD dictionary contains $\tilde{\varepsilon}$ proportion of OOD samples out of $l$ total samples, the calibrated probability functions are defined as:
$\tilde{p}_{i n}(\mathbf{z}_i) = \hat{p}_{i n}(\mathbf{z}_i)-\sigma_{in} \hat{p}_{i n}(\mathbf{z}_i)$,  $\tilde{p}_{out}(\mathbf{z}_i) = \hat{p}_{out}(\mathbf{z}_i)-\sigma_{out} \hat{p}_{out}(\mathbf{z}_i)$, where $ \sigma_{in} = \eta_1(1-\tilde{\varepsilon})\tilde{p}_{cal}(\mathbf{z}_i)$, $ \sigma_{out} = \eta_2\tilde{\varepsilon}\tilde{p}_{cal}(\mathbf{z}_i)$, $\tilde{p}_{cal}(\mathbf{z}_i) =\frac{\tilde{k}}{c_b l\left(\tilde{r}_k\left(\mathbf{z}_i\right)\right)^{m-1}}$, and $\tilde{r}(\mathbf{z}_i)$ is the distance obtained via KNN OOD detection using the OOD dictionary. $\eta_1$, $\eta_2$, $\tilde{k}$ are constants, and $\tilde{\varepsilon}$ is bounded by a function related to $\eta_1$ and $\eta_2$.
\begin{theorem}
Given the setup above, if $\hat{p}_{\text {out }}\left(\mathbf{z}_i\right)=\hat{c}_0 \mathbf{1}\left\{\hat{p}_{\text {in }}\left(\mathbf{z}_i ;\right)<\frac{\beta \varepsilon \hat{c}_0}{(1-\beta)(1-\varepsilon)}\right\}$, $\tilde{\varepsilon}>\frac{\eta_1}{\eta_1+\eta_2}$and $\lambda=$ $-\sqrt[m-1]{\frac{(1-\beta)(1-\varepsilon) k}{\beta \varepsilon c_b n \hat{c}_0}}$, then there exists $\delta > 0$ such that:

\label{theorem1}
\begin{equation}
\mathbf{1}\left\{-r_k\left(\mathbf{z}_i\right) \geq \lambda\right\}=\mathbf{1}\left\{\tilde{p}\left(g_i=1 \mid \mathbf{z}_i\right) \geq \beta + \delta\right\}
\end{equation}

\end{theorem}
\begin{lemma}
With the setup specified above and the premise of Theorem \ref{theorem1}, the decision boundary can be rewritten as:
\label{lemma1}
\end{lemma}
\begin{equation}
    \mathbf{1}\left\{-r_k\left(\mathbf{z}_i\right) \geq \lambda\right\}  =\mathbf{1}\left\{\hat{p}\left(g_i=1 \mid \mathbf{z}_i\right) \geq \beta \right\} \nonumber
\end{equation}
\begin{equation}
     =\mathbf{1}\left\{\frac{k(1-\varepsilon)}{k(1-\varepsilon)+\varepsilon c_b n \hat{p}_{\text {out }}\left(\mathbf{z}_i\right)\left(r_k\left(\mathbf{z}_i\right)\right)^{m-1}} \geq \beta\right\}.
\end{equation}
The proof of Lemma \ref{lemma1} is given in \cite{sun2022out}.

\noindent\textit{Proof.}
The calibrated probability of $\mathbf{z_i}$ being ID data is:
\begin{equation}
    \tilde{p}\left(g_i=1 \mid \mathbf{z}_i\right)=\frac{(1-\varepsilon)\tilde{p}_{in}(\mathbf{z}_i)}{(1-\varepsilon)\tilde{p}_{in}(\mathbf{z}_i)+\varepsilon\tilde{p}_{out}(\mathbf{z}_i)} \nonumber
\end{equation}  
\begin{equation}
=\frac{(1-\varepsilon)\hat{p}_{in}(\mathbf{z}_i)}{(1-\varepsilon)\hat{p}_{in}(\mathbf{z}_i)+\frac{1-\sigma_{out}}{1-\sigma_{in}}\varepsilon\hat{p}_{out}(\mathbf{z}_i)}. 
 \end{equation} 
Since $\tilde{\varepsilon} > \frac{\eta_1}{\eta_1 + \eta_2}$, using $\frac{1-\sigma_{out}}{1-\sigma_{in}}=\frac{1-\eta_2 \tilde{\varepsilon}\tilde{p}_{cal}}{1-\eta_1 (1-\tilde{\varepsilon})\tilde{p}_{cal}}<\frac{1-\eta_2 \frac{\eta_1}{\eta_1+\eta_2}\tilde{p}_{cal}}{1-\eta_1 (1-\frac{\eta_1}{\eta_1+\eta_2})\tilde{p}_{cal}}=1$, and combining with Lemma \ref{lemma1}, we obtain:
\begin{equation}
\tilde{p}\left(g_i=1 \mid \mathbf{z}_i\right)>\frac{(1-\varepsilon)\hat{p}_{in}(\mathbf{z}_i)}{(1-\varepsilon)\hat{p}_{in}(\mathbf{z}_i)+\varepsilon\hat{p}_{out}} \nonumber
\end{equation} 
\begin{equation}
=\frac{k(1-\varepsilon)}{k(1-\varepsilon)+\varepsilon c_b n \hat{p}_{\text {out }}\left(\mathbf{z}_i\right)\left(r_k\left(\mathbf{z}_i\right)\right)^{m-1}} \geq \beta. 
 \end{equation} 
Finally, there exists $\delta > 0$ such that:
\begin{equation}
\vspace{-0.14cm}
    \mathbf{1}\left\{-r_k\left(\mathbf{z}_i\right) \geq \lambda\right\}=\mathbf{1}\left\{\tilde{p}\left(g_i=1 \mid \mathbf{z}_i\right) \geq \beta + \delta\right\} \nonumber
\end{equation}

\end{document}